\documentclass{ws-ijprai}

\usepackage{epsfig}
\usepackage{graphicx}
\usepackage{amsmath}
\usepackage{amssymb}
\usepackage{url}
\usepackage{tabularx}
\usepackage{color}
\usepackage{hyperref}
\usepackage{caption}
\usepackage{subcaption}
\usepackage{multirow}

\begin{document}

\markboth{Trung-Nghia Le and Akihiro Sugimoto}{Region-Based Multiscale Spatiotemporal Saliency for Video}

%
\catchline{}{}{}{}{}
%

\title{Region-Based Multiscale Spatiotemporal Saliency for Video}

\author{Trung-Nghia Le}

\address{Department of Informatics, SOKENDAI (Graduate University for Advanced Studies)\\
2-1-2, Hitotsubashi, Chiyoda-ku, Tokyo 101-8430, Japan\\
\email{ltnghia@nii.ac.jp}}

\author{Akihiro Sugimoto}

\address{National Institute of Informatics\\
2-1-2, Hitotsubashi, Chiyoda-ku, Tokyo 101-8430, Japan\\
\email{sugimoto@nii.ac.jp}}

\maketitle


\begin{abstract}
Detecting salient objects from a video requires exploiting both spatial and temporal knowledge included in the video. We propose a novel region-based multiscale spatiotemporal saliency detection method for videos, where static features and dynamic features computed from the low and middle levels are combined together. Our method utilizes such combined features spatially over each frame and, at the same time, temporally across frames using consistency between consecutive frames. Saliency cues in our method are analyzed through a multiscale segmentation model, and fused across scale levels, yielding to exploring regions efficiently. An adaptive temporal window using motion information is also developed to combine saliency values of consecutive frames in order to keep temporal consistency across frames. Performance evaluation on several popular benchmark datasets validates that our method outperforms existing state-of-the-arts.
\end{abstract}



\section{Introduction}

Visual saliency that reflects sensitivity of human vision aims at locating informative and interesting regions in a scene. It is originally developed to predict human eye fixations on images, and has been recently extended to detect salient objects. Computational methods developed for salient object detection are useful for high-level tasks in computer vision and computer graphics. For instance, these methods have been successfully applied in many areas such as object detection\cite{Guo-Neurocomputing2014}, scene classification\cite{Borji-ICRA2011}
, image and video compression\cite{Chenlei-TIP2010}
, image editing and manipulating\cite{Hagiwara-PETMEI2011}
, and video re-targeting\cite{Taoran-ICIP2010}.

Pixel-based computational methods for saliency\cite{Margolin-CVPR2013}\cite{Zhang-ICIP2013} have been mainly developed and achieved good results for detecting static objects; thus they are popular in detecting salient objects for images. However, these approaches take disadvantages in the context of dynamic scene in videos. Videos usually have lower quality than images due to lossy compression, which makes every pixel value always changes over time regardless it belongs to static objects or dynamic motions. Accordingly, pixel-based saliency detection methods could be misled because of this pixel fluctuation. In contrast, region-based saliency detection methods\cite{Zhou-CVPR2014}\cite{Nghia-PSIVT2015} are more effective in videos because these methods have less fluctuation and can capture dynamics in videos better than pixel-based methods. Therefore, in this work, we focus on analyzing videos at regional levels through superpixel segmentation.

\begin{figure} [t]
    \centering
        \includegraphics[width=0.7\linewidth]{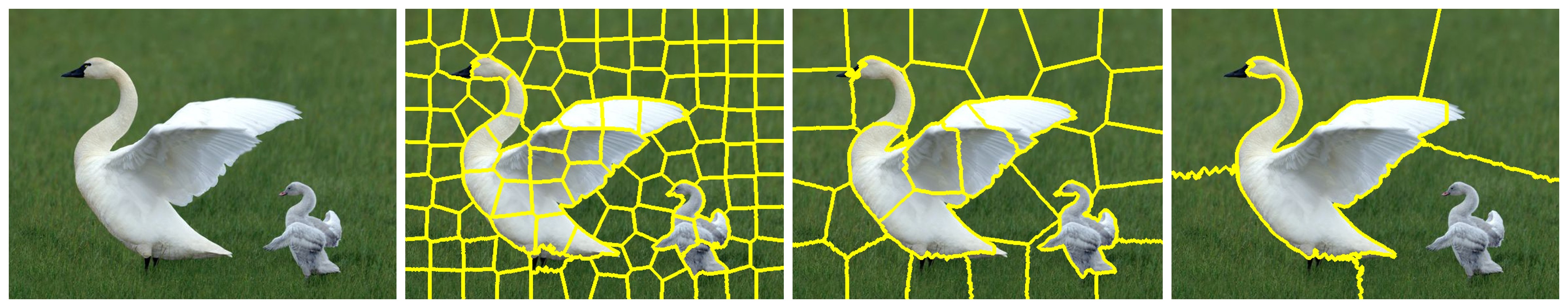}
        \footnotesize
    \caption{Multiscale image analysis. From the left to the right are pixel-wise analysis and region-wise analysis using SLIC superpixel\protect\cite{Achanta-PAMI2012}, followed by small scale superpixels fully covering baby bird and large scale superpixels fully covering mother bird.}
    \label{img:superpixel}
\end{figure}

Superpixels, which are used to detect regions in an image, can be basic materials to capture salient objects at regional levels. Since objects in scenes generally contain various salient scale patterns, superpixel-based regions with a pre-defined size cannot fully explore objects (c.f. Fig.\ref{img:superpixel}). As a result, generated region-based saliency could generally be misled by the complexity of patterns in natural images. This problem can be solved by extending the size of superpixels gradually through the multiscale segmentation approach (c.f. Fig.\ref{img:superpixel}). Multiscale segmentation enables us to analyze saliency cues from multiple scale levels of structure, yielding to dealing with complex salient structures.

Majority of existing methods for saliency, on the other hand, are based on low-level features in scenes such as color, orientation, and intensity 
\cite{Zhang-ICCV2013}\cite{Cheng-ICCV2013}. These bottom-up cues localize objects that present distinct characteristics from their surroundings. However, they do not concern any explicit information about scene structure, context or task-related factors. Therefore, they cannot effectively highlight objects in videos because they do not always reflect regions corresponding to moving parts. In contrast, middle-level features such as objectness\cite{Alexe-CVPR2010} and background prior\cite{Zhu-CVPR2014} are suitable for exploiting moving objects because they focus only on properties in distinct regions. Therefore, combining middle-level features with low-level features can boost up current methods and improve the performance. 
In terms of saliency cues, our method exploits both low-level and middle-level features based on not pixels but regions.

A video is usually composed of dynamic entities caused by egocentric movements or dynamics of the real world. Particularly, in a dynamic scene, background always changes; different parts corresponding to different elements or objects can move in different directions with different speed independently. Saliency models should have ability to fuse current static information and accumulated knowledge on dynamics from the past to deal with the dynamic nature of scenes including two properties: dynamic background and entities' independent motion. Several spatiotemporal saliency detection methods based on motion analysis are proposed for videos\cite{Liu-PAMI2011}\cite{Rahtu-ECCV2010}\cite{Zhai-MM2006}. Some of them can capture scene regions that are important in a spatiotemporal manner\cite{Wang-CVPR2015}\cite{Zhou-CVPR2014}. However, most of existing methods do not fully exploit the nature of dynamics in a scene. Temporal features presenting motion dynamics of objects in a scene between consecutive frames are not utilized in saliency detection process, either.

\begin{figure} [t]
    \centering
        \includegraphics[width=1\linewidth]{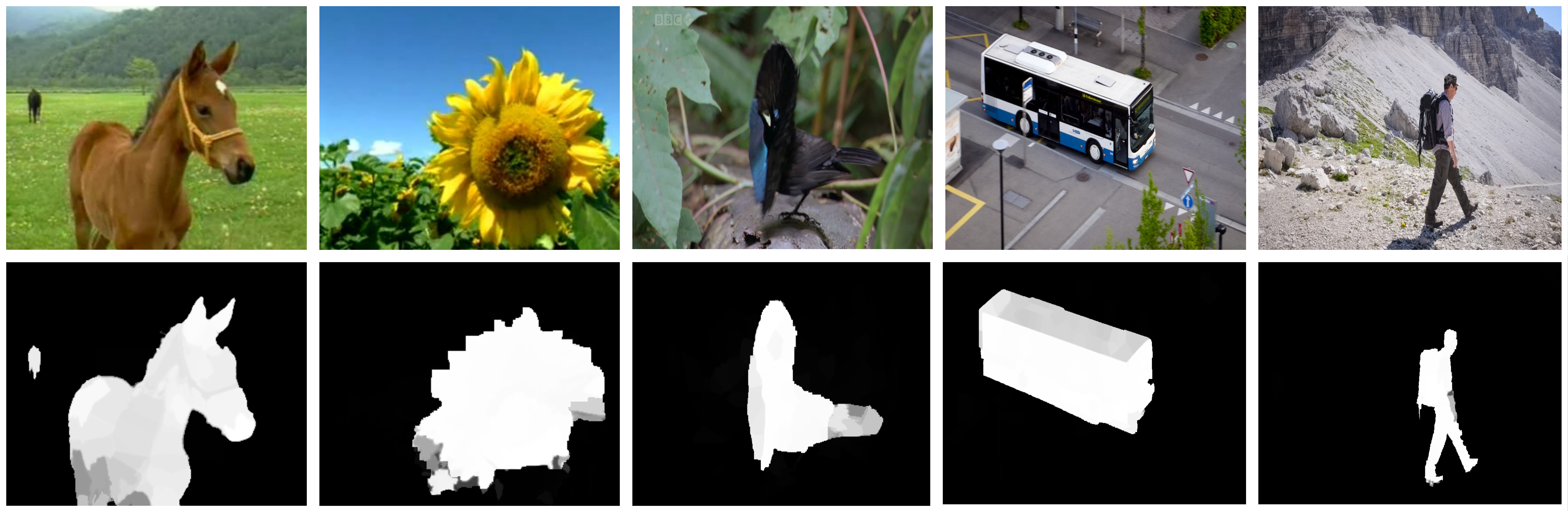}
    \caption{Examples of our spatiotemporal saliency detection method. Top row images are original images. Bottom row images are the corresponding saliency maps using our method.}
    \label{img:examples}
\end{figure}

In order to effectively use knowledge on dynamics of background and objects in a video, we propose a salient object detection method where low-level features 
and middle-level features 
are fused. In this framework, static features and dynamic features computed from the low level and the middle level are combined together to utilize both spatial features of each frame and consistency between consecutive frames (c.f. Table \ref{tab:feature}). The features are exploited in a region-based multiscale saliency model, where saliency cues from multiple scale levels in the structure are analyzed and integrated to take advantage of each level. Using region-based features and multiscale analysis, our method is able to deal with complex scale structures in terms of dynamic scene, so that salient objects are labeled more accurately. We also present a novel metric for motion information by estimating the number of referenced frames for each single object to keep temporal consistency across frames. Our method overcomes the limitation of the existing method\cite{Zhou-CVPR2014} which uses a fixed number of referenced frames and does not concern motion of objects within a scene. Examples of generated saliency maps using our method are shown in Fig. \ref{img:examples}.


Our key contributions lie in twofold:

\begin{itemlist}
\item We propose a region-based multiscale framework, which explicitly integrates low-level features together with middle-level features. Our regional features are exploited to analyze saliency cues from multiple scale levels of structure. With using region-based features and multiscale analysis, our method is able to deal with complex scale structures in terms of dynamic scenes, so that salient objects are labeled more accurately. Although the proposed saliency model is developed based on the framework presented by Zhou et al.\cite{Zhou-CVPR2014}, it significantly improves the performance of the original work.

\vspace*{0.3\baselineskip}

\item We introduce a novel metric called adaptive temporal window using motion information in order to keep temporal consistency between consecutive frames of each entity in a video. Our method also exploits the dynamic nature of the scene in term of independent motion of entities.
\end{itemlist}

The rest of this paper is organized as follows. In Section \ref{section:related_work}, we briefly present and analyze the related work in saliency models for videos as well as region segmentation in videos. The proposed method is presented in Section \ref{section:proposed_method}. Experiments are showed in Section \ref{section:experimental_setup}, Section \ref{section:validation}, and Section \ref{section:comparison}. Finally, Section \ref{section:conclusion} presents conclusion and ideas for future work. We remark that a part of this work has been reported in \cite{Nghia-PSIVT2015}.

\begin{table}[t]
\centering
\caption{Our used feature classification.}
\begin{tabular}{|l|l|l|}
\hline
                          & \multicolumn{1}{c|}{\textbf{\begin{tabular}[c]{@{}c@{}}low-level feature\end{tabular}}} & \multicolumn{1}{c|}{\textbf{\begin{tabular}[c]{@{}c@{}}middle-level feature\end{tabular}}} \\ \hline
\textbf{static feature}   & \begin{tabular}[c]{@{}l@{}}- color\\ - intensity\\ - orientation\end{tabular}                & \begin{tabular}[c]{@{}l@{}}- center bias\\ - objectness\cite{Alexe-CVPR2010}\\ - background prior\cite{Zhu-CVPR2014}\end{tabular}                             \\ \hline
\textbf{dynamic feature} & \begin{tabular}[c]{@{}l@{}}- flow magnitude\\ - flow orientation\end{tabular}             & \ - movement                                                                                   \\ \hline
\end{tabular}
\label{tab:feature}
\end{table}


\section{Related Work}\label{section:related_work}
In this section, we briefly review related work in saliency models for videos and region segmentation in videos.

\subsection{Dynamics in saliency detection for videos}

The spatial saliency models for still images can be frame-wisely applied to videos to detect salient objects. 
Some techniques such as fast minimum barrier distance transform\cite{Zhang-ICCV2015} and minimum spanning tree\cite{Tu-CVPR2016} are used to develop real-time salient object detection systems, which can apply to videos. Other methods formulate the problem based on boolean map theory\cite{Zhang-ICCV2013}, absorbing Markov chain\cite{Jiang-ICCV2013}, and weighted sparse coding framework\cite{Nianyi-CVPR2015}. Multiple saliency maps can be aggregated by conditional random field\cite{Liu-PAMI2011} or cellular automata dynamic evolution model\cite{Quin-CVPR2015}. However, this approach does not achieve high effectiveness because they cannot exploit temporal knowledge in dynamic scenes in videos.

Dynamic cues have usually been employed for saliency detection to deal with dynamics in videos. Some video saliency models\cite{Seo-JOV2009}\cite{Zaharescu-ACCV2012} reply on center-surround differences between a local spatiotemporal cube and its neighboring cubes in space-time coordinates to find salient regions in dynamic image streams. L.Itti et al.\cite{Itti-CVPR2005} developed a method which computes instantaneous low-level surprise at every location in complex video clips. Flicker is added into saliency detection as a new cue to build a neurobiological model of visual attention for automatic realistic generation of eye and head movements given in video scenes\cite{Itti-AMOST2004}. In the method proposed by M.Mancas et al.\cite{Mancas-ICIP2011}, only dynamic features such as speed and direction are used to quantify rare and abnormal motion.  

Motion cues, which are considered as a low-level feature channel, have recently been employed in spatiotemporal frameworks together with spatial features such as illumination, color, and so on. Several spatiotemporal saliency detection methods based on motion analysis are proposed for videos. Motion between a pair of frames, which is considered as optical flow, is used to compute local discrimination of the flow in a spatial neighborhood\cite{Nghia-PSIVT2015}\cite{Zhou-CVPR2014}. Motion features such as optical flow\cite{Rahtu-ECCV2010} or SIFT flow\cite{Liu-PAMI2011} are incorporated into conditional random field to extend the models for still images to detect salient objects from videos. In the spatiotemporal attention detection framework proposed by Y.Zhai et al.\cite{Zhai-MM2006}, motion contrast between consecutive frames is estimated by applying RANSAC algorithm on point correspondences in the scene. W.Wang et al.\cite{Wang-TIP2015} estimated salient regions in videos based on the gradient flow field, which consists of intra-frame boundary and inter-frame motion, and the energy optimization. In the framework proposed by W.Wang et al.\cite{Wang-CVPR2015}, the spatiotemporal saliency map, which is computed from temporal motion boundaries and spatial edges, is combined with the appearance model and the dynamic location model to segment salient video objects. C.Feichtenhofer et al.\cite{Feichtenhofer-CVPR2015} defined a space-time saliency model, which relies on two general observations regarding actions (e.g. motion contrast and motion variance), for capturing foreground action motion. Y.Lou et al.\cite{Lou-ACCV2014} measured relationships within and between trajectories of superpixels over time to capture sudden or onset movements in a scene. H.Kim et al. \cite{Hansang-TIP2015} incorporated the spatial transition matrix and the temporal restarting distribution, which is computed from motion distinctiveness, temporal consistency, and abrupt change, into the random walk with restart framework to detect spatiotemporal saliency. 

However, most of existing methods do not fully exploit the nature of dynamics in a scene. Temporal features presenting motion dynamics of objects in a scene between consecutive frames are not utilized in saliency detection process, either. Differently from existing methods, our spatiotemporal saliency detection method uses motion information to keep temporal consistency across frames. We propose an adaptive temporal sliding window to relate salient values of frame sequences by exploiting motion information of an entity in each frame.

\subsection{Region segmentation in videos}

Most video analysis methods are extended from image analysis methods \cite{Chenliang-ECCV2012}\cite{Chang-CVPR2013}. Although frame by frame processing is efficient and can achieve high performance in spatial respects, its temporal stability is limited. Therefore, many video segmentation methods are developed to exploit the temporal knowledge in videos.  

Superpixel based algorithms are widely used as a pre-processing step in both still images and videos to generate supporting regions and to speed up further computations\cite{Yan-CVPR2013}\cite{Nghia-PSIVT2015}
. Existing superpixel methods can be divided into two categories: one is to grow superpixels starting from an initial set, and the others are based on graphs\cite{Bergh-ECCV2012}.

For the first approach, one of the most efficient superpixel segmentation algorithms for images was introduced by R.Achanta et al.\cite{Achanta-PAMI2012} and is called Simple Linear Iterative Clustering (SLIC). In this algorithm, the $k$-means clustering is firstly performed to group pixels that exhibit similar appearances into superpixels and then single-pixel superpixels are merged into large superpixels via a single 4-connected region. Many segmentation methods have been proposed recently based on the SLIC superpixel. J.Chang et al.\cite{Chang-CVPR2013} extended SLIC to propose a probabilistic model for temporally consistent superpixels in video sequences. SEED superpixel method\cite{Bergh-ECCV2012} also shares the idea of growing superpixels from an initial set with SLIC. However, the SEED superpixel directly exchanges pixels between superpixels by moving boundaries instead of growing superpixels by clustering pixels around the centers. From the SEED superpixel, M.Van den Bergh et al.\cite{Bergh-ICCV2013} proposed an online, real-time video segmentation algorithm to exploit temporal information in videos.


In addition, there are some segmentation methods based on graphs. 
C.Xu et al.\cite{Chenliang-ECCV2012} developed a framework for streaming video sequences based on hierarchical graph-based segmentation. Using a spatiotemporal extension of GraphCut method\cite{Lee-ICCV2011}, A.Papazoglou et al.\cite{Papazoglou-ICCV2013} introduced a fast segmentation method for unconstrained videos which include rapidly moving background, arbitrary object motion and appearance, as well as non-rigid deformations and articulations.

In this work, we employed the temporal superpixel method proposed by J.Chang\cite{Chang-CVPR2013} in our framework. This is because it overcomes limitations of other video segmentation methods and is popularly employed in the computer vision literature\cite{Ros-WACV2015}\cite{Kae-CVPR2014}.


\section{Multiscale Spatiotemporal Saliency Detection Method}\label{section:proposed_method}
\subsection{Overview}
\begin{figure*}[t]
    \captionsetup{justification=centering}
    \centering
        \includegraphics[width=1\textwidth]{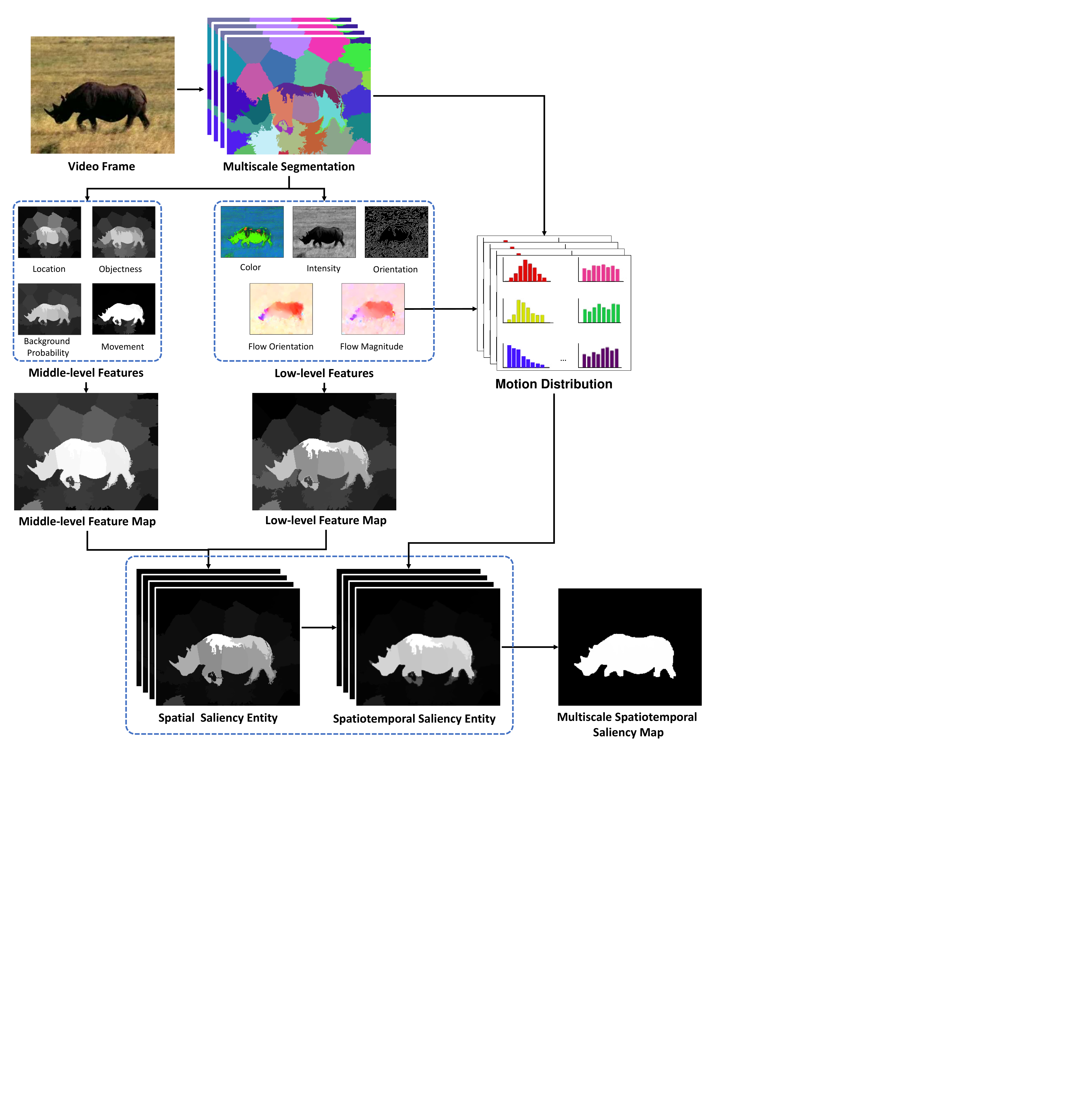}
    \caption{Pipeline of the proposed spatiotemporal saliency detection method.}
    \label{fig:overview}
\end{figure*}

The goal of this work is to detect salient regions in videos by combining static features with dynamic features where the features are detected from regions but not from pixels. Figure \ref{fig:overview} illustrates the process of our multiscale spatiotemporal saliency detection method. 

First of all, the temporal superpixels model\cite{Chang-CVPR2013} is executed to segment a video into spatiotemporal regions at various scale levels. Motion information, as well as features for each frame, are extracted at each scale level. From these features, we build feature maps, including both low-level feature maps presenting contrasts between regions and middle-level feature maps presenting properties inside regions. These two kinds of feature maps are combined to generate spatial saliency entities for regions at each scale level. Temporal consistency is incorporated into spatial saliency entities to form spatiotemporal saliency entities by using an Adaptive Temporal Window (ATW) for each region individually to smooth saliency values across frames. Finally, a spatiotemporal saliency map is generated for each frame by fusing its multiscale spatiotemporal saliency entities.

\subsection{Multiscale video segmentation}

To support the intuition that objects in a video generally contain various salient scale patterns as well as an object at a coarser scale may be composed of multiple parts at a finer scale, the video is segmented at multiple scales. Multiscale segmentation enables us to analyze saliency cues from multiple scale levels of structure, yielding to dealing with complex salient structures (c.f. Fig. \ref{img:superpixel}). In this work, we segment a video at three scale levels. We also remark that each segmentation level has a different number of superpixels, which are defined as non-overlapping regions.

To segment a video, we employed the temporal superpixel method\cite{Chang-CVPR2013}, which is based on multiple frame superpixel segmentation. This method is extended from SLIC\cite{Achanta-PAMI2012}. Differently from the SLIC method, the temporal superpixel method utilizes a spatial intensity Gaussian Mixture Model (GMM) combined with a motion model, served as a prior for the next frame. Motion information is used to propagate superpixels over frames to reduce generative superpixels in a single frame.

After the segmentation process, we obtain multiple scale temporal superpixels. At each scale, superpixels across frames are connected, thus we can predict motion of a superpixel over frames through its positions at frames.

\subsection{Spatial saliency entity construction}
\subsubsection{Low-level feature map}\label{section:pixelwise_feature_map}
Human vision reacts to image regions with discriminative features such as unique color, high contrast, different orientation, or complex texture\cite{Nataraju-ML11}. To estimate attractiveness of regions in a video, contrast metric is usually used to evaluate the sensitivity of elements in each frame. The contrast is usually based on low-level features including static information such as color, intensity, or texture, and dynamic information such as magnitude or orientation of motion. A region with high contrast against surrounding regions can attract human attention and is perceptually more important.

For the $i$-th region at the $l$-th scale of the segmentation model at a frame, denoted by $r_{i,l}$, we compute its normalized color histogram in the CIE Lab color space, denoted by $\chi _{i,l}^{{lf_{col}}}$, and distribution of lightness $\chi _{i,l}^{{lf_{lig}}}$. We quantize the four color channels (L, A, B and Hue) into 16 bins for each channel to compute $\chi _{i,l}^{{lf_{col}}}$. We also uniformly quantize $\chi _{i,l}^{{lf_{lig}}}$ into 16 bins.

We also calculate orientation statistic $\chi _{i,l}^{{lf_{ori}}}$ of the region $r_{i,l}$. We use the following 2-D Gabor functions\cite{Daugman-JOSA85} to model the image texture for every pixel $\left(x,y\right)$:
\begin{equation}
\begin{array}{l}
{g_{\lambda ,\varphi ,\gamma ,\sigma ,\theta }}\left( {x,y} \right) = \exp \left( { - \frac{{x{'^2} + {\gamma ^2}y{'^2}}}{{2{\sigma ^2}}}} \right)\cos \left( {2\pi \frac{{y'}}{\lambda } + \varphi } \right) ,\\
x' = x\cos \theta  + y\sin \theta , \\
y' =  - x\sin \theta  + y\cos \theta , 
\end{array}
\label{equation:gabor}
\end{equation}
where $\gamma, \lambda ,\sigma, \varphi,$ and $\theta$ are parameters as follows: 
$\gamma=0.5$ is a constant, called the spatial aspect ratio, which specifies the ellipticity of the support of the Gabor function. $\lambda=8$ denotes the wavelength of the cosine factor of the Gabor filter kernel and herewith the preferred wavelength of the Gabor function.  
$\sigma$ is the standard deviation of the Gaussian factor of the Gabor function where the ratio $\sigma/\lambda$ determines the spatial frequency bandwidth. In this paper, we fix $\sigma=0.56\lambda$ corresponding to a bandwidth of one octave at half-response, corresponding to $bw=1$ in $\sigma  = \frac{{{2^{bw}} + 1}}{{{2^{bw}} - 1}} \cdot \frac{\lambda }{\pi }\sqrt {\frac{1}{2}\log 2}$. 
$\varphi$ is a phase offset that determines the symmetry filter of the Gabor function. We use quadrature pairs of two filter banks, including an odd filter with $\varphi=\pi$ and an even filter with $\varphi=0$.  
The angle parameter $\theta$ specifies the orientation of the normal to the parallel stripes of the Gabor function. In this work, we use 8 orientations $\theta  = k\frac{{\pi }}{4}$ where $k \in \left\{ {0,1,...,7} \right\}$. We then quantize $\chi _{i,l}^{{lf_{ori}}}$ into 16 bins.

Since the human visual system is more sensitive to moving objects than still objects, dynamic features are also compared between regions at the same segmentation level\cite{Hillstrom-PP94}. Pixel-wise optical flow\cite{Liu-ThesisMIT2009} is used to analyze motion between consecutive frames. Regional motion features of a region are obtained by computing distribution of this flow information in the region. The motion distribution of region $r_{i,l}$ is encoded in two descriptors: $\chi _{i,l}^{{lf_{fmag}}}$ is a normalized distribution of flow magnitude and $\chi _{i,l}^{{lf_{fori}}}$ is a normalized histogram of flow orientation. We uniformly quantize $\chi _{i,l}^{{lf_{fmag}}}$ and $\chi _{i,l}^{{lf_{fori}}}$ into 16 and 9 bins respectively.

The low-level feature map of each region is considered as the sum of its feature distances to other neighbor regions at the same scale level in the segmentation model with different weight factors:
\begin{equation}
{S_{{lf}_{i,l}}} = \sum\limits_{lf} {{w_{lf}}} \sum\limits_{j \ne i} {\left| {{r_{j,l}}} \right|\omega \left( {{r_{i,l}},{r_{j,l}}} \right)\left\| {\chi _{i,l}^{lf} - \chi _{j,l}^{lf}} \right\|},
\label{equation:contrast}
\end{equation}
where $\left\| {\chi _{i,l}^{lf} - \chi _{j,l}^{lf}} \right\|$ is the Chi-Square distance\cite{Pele-ECCV2010} between two histograms, $lf \in \left\{ {{lf_{col}},\,{lf_{lig}},\,{lf_{ori}},\,{lf_{fmag}},\,{lf_{fori}}} \right\}$ denotes one of the five features with corresponding weight $w_{lf} \in \left\{ {0.4, 0.1, 0.1, 0.2, 0.2} \right\}$. $\left| {{r_{j,l}}} \right|$ denotes the contrast weight of region $r_{j,l}$, which is the ratio of its size to the frame size. Regions with more pixels contribute higher contrast weight factors, than those containing smaller number of pixels. $\omega\left( {{r_{i,l}},{r_{j,l}}} \right)$ controls spatial distance influence between two regions $r_{i,l}$ and $r_{j,l}$:
\begin{equation}
\omega \left( {{r_{i,l}},{r_{j,l}}} \right) = {e^{\frac{{ - D{{\left( {{r_{i,l}},{r_{j,l}}} \right)}^2}}}{{{\sigma_{sp-dst}^2}}}}},
\label{equation:distance_contrast}
\end{equation}
where $D{{\left( {{r_{i,l}},{r_{j,l}}} \right)}}$ is the Euclidean distance between region centers and parameter $\sigma_{sp-dst}=0.2$ controls how large the neighbors are. Finally, ${S_{{lf}_{i,l}}}$ is normalized to range $\left[ {0,1} \right]$.

\subsubsection{Middle-level feature map}\label{section:regionwise_feature_map}

In addition to contrasts between regions, we also compute properties of each region based on middle-level features. We observe that human vision is biased toward specific spatial information of a video such as the center of the frame, foreground objects, or background as well as movements of objects over time. Therefore, our middle-level features are based on center bias, objectness, background prior and movement metrics.

Human eye-tracking studies show that human attention favors the center of natural scenes when watching videos\cite{TSeng-JOV2009}. So, pixels close to the screen center could be salient in many cases. Our center bias is defined as:
\begin{equation}
{\chi _{i,l}}^{{{mf}_{cen}}} = \frac{1}{{\left| {{r_{i,l}}} \right|}}\sum\limits_{j \in {r_{i,l}}} {{e^{\frac{{ - D{{\left( {{p_j},\bar p} \right)}^2}}}{{{\sigma_{cen}^2}}}}}},
\label{equation:distance_center}
\end{equation}
where $\left| {{r_{i,l}}} \right|$ denotes the number of pixels in region ${r_{i,l}}$ and $D\left( {{p_j},\bar p} \right)$ is the Euclidean distance from each pixel $p_j$ in the region to the image center $\bar p$. $\sigma_{cen}=0.3$ is the parameter.

The second characteristic is objectness, which is a class-generic object detector\cite{Alexe-CVPR2010}. It quantifies how likely it is for an image window to contain an object of any class. Our objectness feature is defined as:
\begin{equation}
\chi _{i,l}^{{f_{obj}}} = \frac{1}{{\left| {{r_{i,l}}} \right|}}\sum\limits_{j \in {r_{i,l}}} {{o_j}},
\end{equation}
where ${o_j}$ is pixel-wise objectness map in region $r_{i,l}$. Objectness map provides meaningful distributions over the object locations, demonstrating probability of containing objects at pixels. Four object cues are utilized as follows: multi-scale saliency $MS$ measuring uniqueness characteristic of objects, color contrast $CC$ measuring different appearance from their surroundings, edge density $ED$ and superpixel straddling $SS$ measuring closed boundary characteristic of objects. These object cues are combined into a Bayesian framework:
\begin{equation}
o_j=p^{(j)}\left( {obj|A} \right) = \frac{{p^{(j)}\left( {A|obj} \right)p^{(j)}\left( {obj} \right)}}{{p^{(j)}\left( A \right)}} = \frac{{p^{(j)}\left( {obj} \right)\prod\limits_{cue \in A} {p^{(j)}\left( {cue|obj} \right)} }}{{\sum\limits_{c \in \left\{ {obj,bgr} \right\}} {p^{(j)}\left( c \right)\prod\limits_{cue \in A} {p^{(j)}\left( {cue|c} \right)} } }},
\end{equation}
where $A=\left\{ {MS,CC,ED,SS} \right\}$ denotes object cues, and $p^{(j)}\left( \cdot \right)$ is the probability at pixel $j$. Priors $p\left(obj\right)$, $p\left(bg\right)$, individual cue likelihoods $p\left(cue|c\right)$ and $c \in \left\{ {obj,bg} \right\}$, are estimated from the training dataset. This posterior constitutes the objectness score of overlapping regions.

The regional background prior is based on differences of spatial layout of image regions\cite{Zhu-CVPR2014}. Object regions are much less connected to image boundaries than background ones. In contrast, a region corresponding to background tends to be heavily connected to the image boundary. In order to compute background probability of each region, each segmented image is built as an undirected weighted graph by connecting all adjacent regions and assigning their weights $w_{edge}$ as the Euclidean distance between their average colors in the CIE-Lab space. The background feature of region $r_{i,l}$ is written as:
\begin{equation}
{\chi _{i,l}}^{{{mf}_{bgr}}} = \exp \left( { - \frac{{BndCo{n^2}\left( {{r_{i,l}}} \right)}}{{2\sigma _{bgr}^2}}} \right),
\end{equation}
where $BndCon\left( {{r_{i,l}}} \right)$ is the boundary connectivity of region $r_{i,l}$. Similarly to\cite{Zhu-CVPR2014}, we set the parameter $\sigma _{bgr}=1$. The boundary connectivity of region $r_{i,l}$ is calculated as the ratio of length along the boundary of it to the square root of its spanning area:
\begin{equation}
{\mathop{\rm BndCon}\nolimits} \left( {{r_{i,l}}} \right) = \frac{{{{{\mathop{\rm len}\nolimits} }_{bnd}}\left( {{r_{i,l}}} \right)}}{{\sqrt {{\mathop{\rm SpanArea}\nolimits} \left( {{r_{i,l}}} \right)} }}
\end{equation}
where ${{{{\mathop{\rm len}\nolimits} }_{bnd}}\left( {{r_{i,l}}} \right)}$ is length along the boundary of region $r_{i,l}$, which is calculated as the sum of the Geodesic distance from it to regions on the boundary at the same scale level. ${{\mathop{\rm SpanArea}\nolimits} \left( {{r_{i,l}}} \right)}$ is the spanning area of region $r_{i,l}$ calculated as the sum of the Geodesic distances from it to all regions in the frame at the same scale level. The Geodesic distance\cite{Zhu-CVPR2014} between any two regions is defined as the accumulated edge weight along their shortest path on the graph:
\begin{equation}
{d_{geo}}\left( {{r_{i,l}},{r_{j,l}}} \right) = \mathop {\min }\limits_{{r_{1,l}} = {r_{i,l}},{r_{2,l}},...,{r_{n,l}} = {r_{j,l}}} \sum\limits_{k = 1}^{n - 1} {{w_{edge}}\left( {{r_{k,l}},{r_{k + 1,l}}} \right)}
\end{equation}

Moreover, to encode movement of objects, we capture any sudden speed change in motion of regions. Movement of a region is calculated as its average motion magnitude values computed from the optical flow\cite{Liu-ThesisMIT2009}:
\begin{equation}
\chi _{i,l}^{{f_{mov}}} = \frac{1}{{\left| {{r_{i,l}}} \right|}}\sum\limits_{j \in {r_{i,l}}} {{m_j}},
\end{equation}
where ${m_j}$ is motion magnitude at all pixels in region $r_{i,l}$.

The middle-level feature map of the region $r_{i,l}$ is computed as the sum of its attribute values with different weight factors:
\begin{equation}
{S_{{mf}_{i,l}}} = \sum\limits_{{mf}} {{w_{{mf}}}{\chi}_{i,l}^{{mf}}},
\label{equation:heuristics}
\end{equation}
where ${mf} \in \left\{ {{{mf}_{cen}},\,{{mf}_{obj}},\,{{mf}_{bgr}},\,{{mf}_{mov}}} \right\}$ denotes one of the four features, with corresponding weight factor $w_{mf} \in \left\{ {0.15, 0.05, 0.4, 0.4} \right\}$. Finally, ${S_{{mf}_{i,l}}}$ is normalized to range $\left[ {0,1} \right]$.

\subsubsection{Feature map combination}\label{section:feature_combination}

\begin{figure}[t]
    \centering
    \includegraphics[width=1\linewidth]{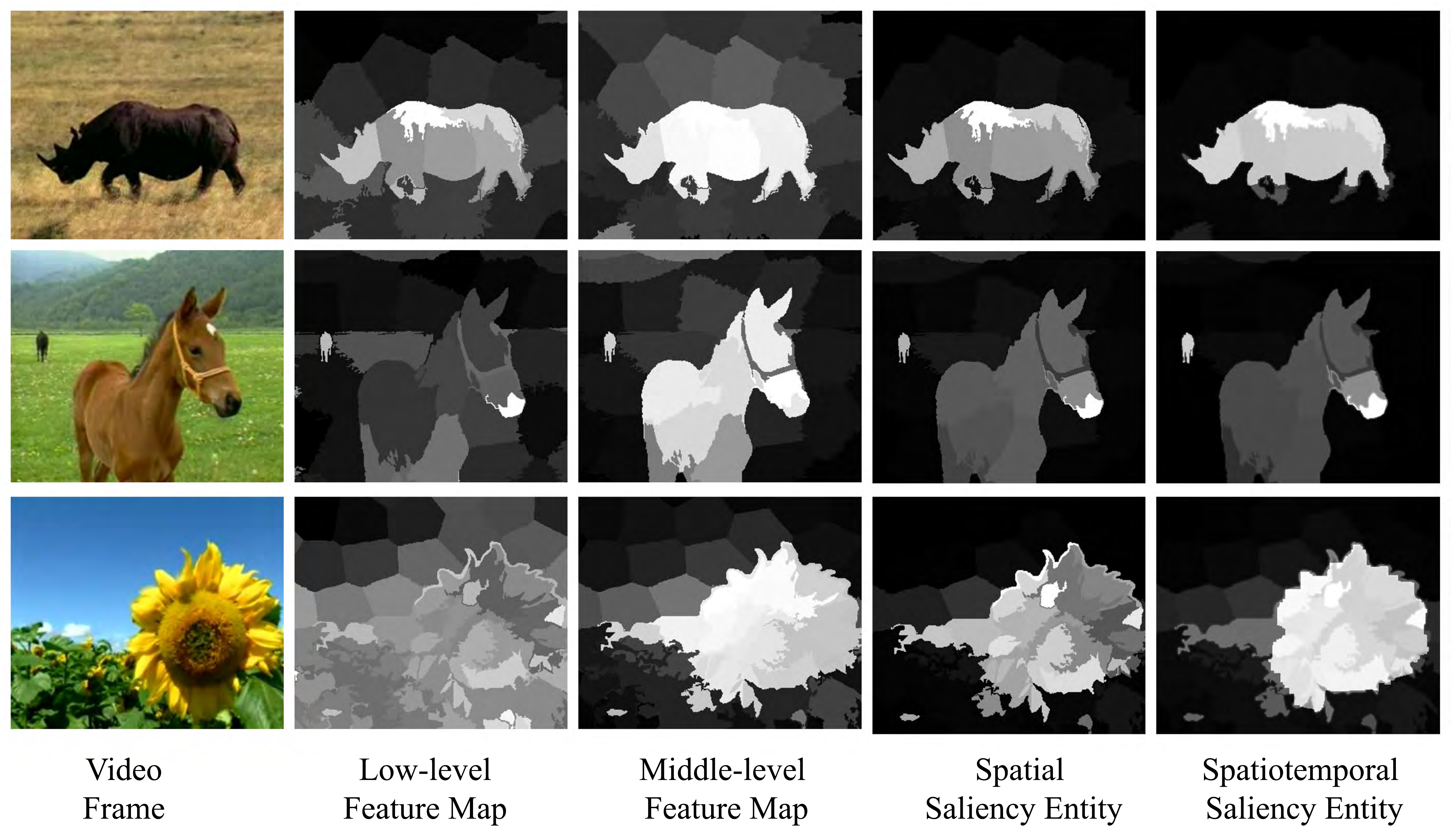}
    \caption{Saliency entity construction.}
\label{img:saliency_entity}
\end{figure}

Combining the low-level feature maps and the middle-level feature maps allows us to obtain initial spatial saliency entities for all segmentation levels separately by a weighted multiplicative integration (c.f. Fig.\ref{img:saliency_entity}):
\begin{equation}
{S_{i,l}} = {S_{{lf}_{i,l}}^\alpha}{S_{{mf}_{i,l}}^{1-\alpha}},
\end{equation}
where the parameter $\alpha$ controls the trade-off between the low-level feature maps and the middle-level feature maps. To weight the low-level feature map and the middle-level feature map equally, we set $\alpha=0.5$. Finally, the spatial saliency entities are linearly normalized to the fixed range $\left[ {0,1} \right]$  in order to guarantee that regions with value 1 are the maxima of saliency. We note that we demonstrate effects of using the low-level feature maps and the middle-level feature maps on saliency results in Section \ref{section:validation_combination}.


\subsection{Incorporating temporal consistency}\label{section:ATW}

\begin{figure}[t]
    \centering
        \includegraphics[width=0.8\linewidth]{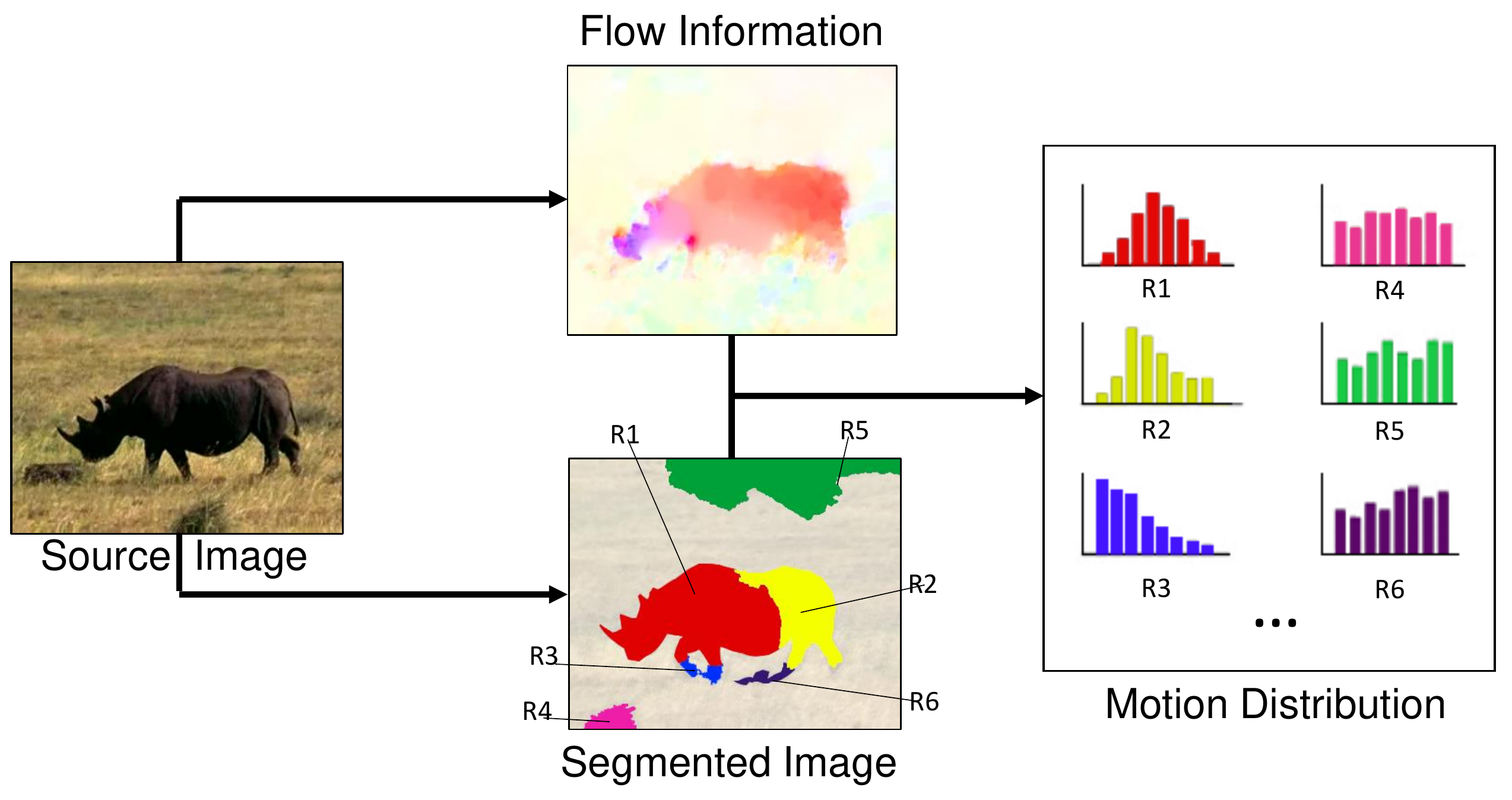}
    \caption{Regional motion calculation.}
    \label{fig:motion_process}
\end{figure}

In a video, it is sometimes hard to distinguish objects from background because every pixel value always changes over time regardless that it belongs to an object or background. Moreover, motion analysis shows that different parts of objects move with various speed and, furthermore, background motion also changes with different speed and direction (c.f. flow information in Fig. \ref{fig:motion_process}).
This causes fluctuation of object appearances between frames. To reduce this negative effect, each spatial saliency entity at the current frame is combined with those at neighboring frames, resulting in smoothing saliency values over time. After this operation, salient values on contiguous frames can become similar, and this generates robust spatiotemporal saliency entities. 

We propose to adaptively use a sliding window in the temporal domain, Adaptive Temporal Window (ATW), for each region at each frame to capture speed variation by exploiting motion information in the region. A spatial saliency entity at each scale level at the current frame is combined with spatial saliency entities at neighboring frames using the Gaussian combination weights, where nearer frames have larger weights:
\begin{equation}
\tilde S_{i,l}^t = \frac{1}{\Psi }\sum\limits_{t' = t - \Phi _{i,l}^t}^t {{e^{\frac{{ - D{{\left( {t,t'} \right)}^2}}}{{2\Phi {{_{i,l}^t}^2}{\sigma_{tp-dst} ^2}}}}}S_{i,l}^{t'}},
\label{equation:ATW}
\end{equation}    
where $\Psi$ is the normalization factor and $S_{i,l}^t$ measures spatial saliency entity of region $r_{i,l}$ at frame $t$ with corresponding weight ${e^{\frac{{ - D{{\left( {t,t'} \right)}^2}}}{{2\Phi {{_{i,l}^t}^2}{\sigma_{tp-dst} ^2}}}}}$. $D{\left( {t,t'} \right)}$ denotes the time difference between two frames, and parameter $\sigma_{tp-dst}=10$ controls how large the region at previous frames is. $\Phi _{i,l}^t$ controls the number of participating frames in the operation, expressed as:
\begin{equation}
\Phi _{i,l}^t = M{e^{ - \mu _{i,l}^t\frac{\lambda }{{\beta _{i,l}^t}}}},
\label{equation:WS}
\end{equation}    
where $M=10$ and $\lambda=2$ are parameters. ${\beta _{i,l}^t} = \frac{{{\sigma _{i,l}^t}}}{{{\mu _{i,l}^t}}}$ is the coefficient variation measuring dispersion of motion distribution of each region. $\mu _{i,l}^t$ and $\sigma _{i,l}^t$ are the mean value and the standard deviation of the motion distribution of region $r_{i,l}$ at frame $t$. To calculate regional motion distribution, we first use the pixel-wise optical flow\cite{Liu-ThesisMIT2009} to compute motion magnitude of each pixel in a frame, and then exploit distribution of motion magnitude in each region (c.f. Figure \ref{fig:motion_process}).

\subsection{Spatiotemporal saliency generation}\label{section:saliency_fusion}

\begin{table}[t]
\centering
\caption{Multi-layer Cellular Automata integration algorithm\protect\cite{Quin-CVPR2015}.}
\label{tab:MCA}
\begin{tabular}{l}
\hline \hline
\textbf{Input:}    $S_l^{t}$ with $l \in \left\{ {1,...,L} \right\}$                            \\
\textbf{Output:} $S{M^t}$                                \\ \hline
\textbf{$\triangleright$ Saliency map binarization by the Otsu algorithm\cite{Ostu-TSMC79}:}                                           \\
\ \ \ \ \ \ $\gamma _l^t \leftarrow \log \left( {\frac{{{\Theta _{ostu}}\left( {S_l^t} \right)}}{{1 - {\Theta _{ostu}}\left( {S_l^t} \right)}}} \right)$ \\
\textbf{$\triangleright$ Saliency refinement:}                                                                   \\
\ \ \ \ \ \ \textbf{for} $k=1 \to K-1$ \textbf{do}                                                                                                                                                         \\ 
\ \ \ \ \ \ \ \ \ \ \ \ $\Lambda \left( {S_l^{t,k + 1}} \right) \leftarrow \Lambda \left( {S_l^{t,k}} \right) + \ln \left( {\frac{\lambda }{{1 - \lambda }}} \right)\sum\limits_{i = 1,i \ne l}^L {{\mathop{\rm sign}\nolimits} \left( {\Lambda \left( {S_i^{t,k}} \right) - \gamma _i^t} \right)}$,     \\
\ \ \ \ \ \ \ \ \ \ \ \ where $\Lambda \left( {S_l^t} \right) = \frac{{S_l^t}}{{1 - S_l^t}}$                                      \\
\ \ \ \ \ \ \textbf{end for}                                                                                                                                               \\
\textbf{$\triangleright$ Saliency map integration:}                                                              \\
\ \ \ \ \ \ $S{M^t} \leftarrow \frac{1}{L}\sum\limits_{l = 1}^L {S_l^t}$                                    \\ \hline
\textbf{Parameter settings\cite{Quin-CVPR2015}:}                                    \\ 
\ \ \ \ \ \ $K=5$ and $\ln \left( {\frac{\lambda }{{1 - \lambda }}} \right) = 0.15$\\ \hline \hline
\end{tabular}
\end{table}

\begin{figure}[t]
    \centering
    \includegraphics[width=1\linewidth]{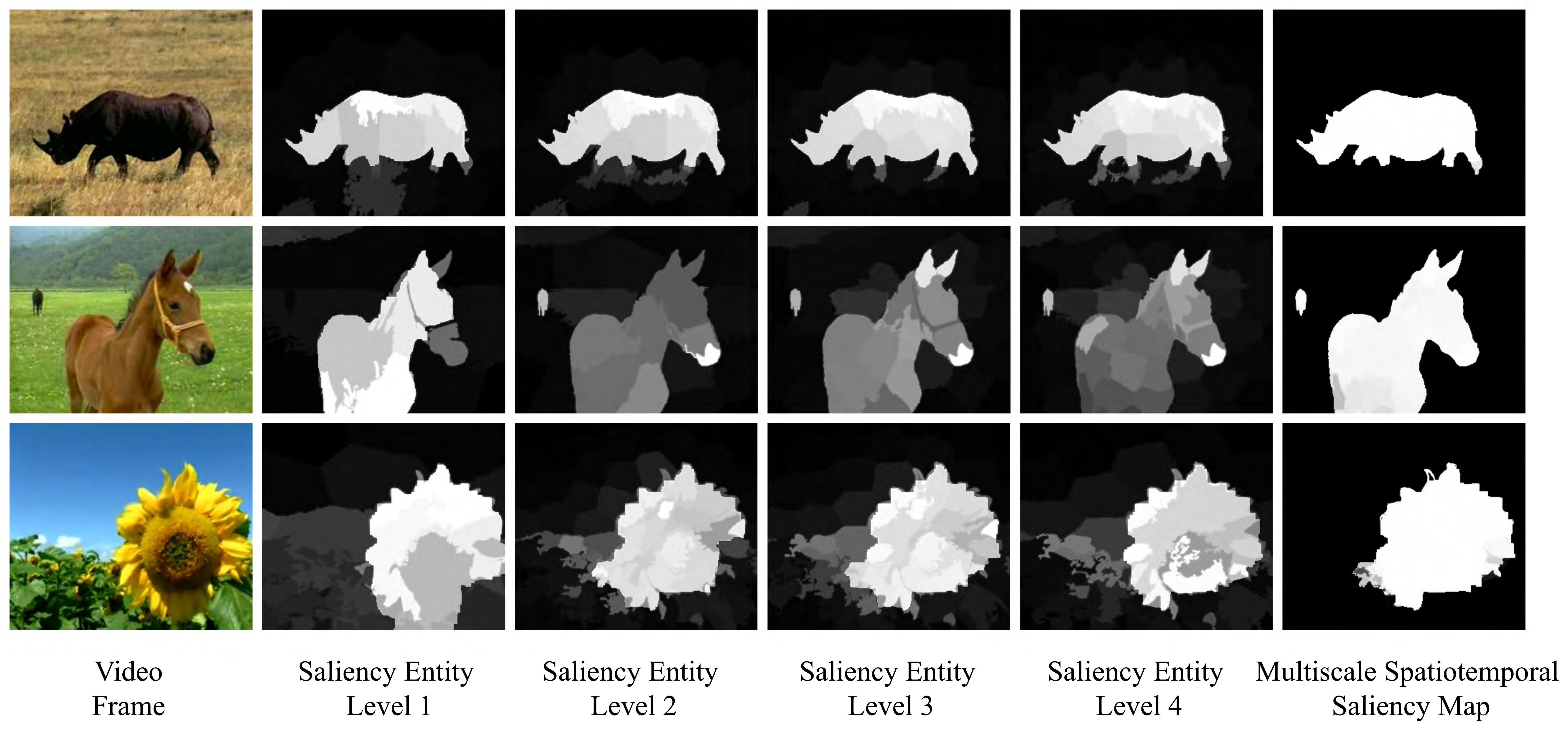}
    \caption{Saliency integration using Multi-layer Cellular Automata\protect\cite{Quin-CVPR2015}.}
\label{img:MCA}
\end{figure}

A multiscale saliency map is usually generated by calculating the average of saliency maps over all segmentation levels. However, this kind of process does not achieve good results because saliency values at a scale level have their own advantages and disadvantages. Compared with naive averaging of all levels, the Multi-layer Cellular Automata (MCA) integration algorithm aggregates confident saliency values from these levels, surely yielding better performance\cite{Quin-CVPR2015}. Therefore, we implement the MCA algorithm to take advantage of multiple scale levels. The generated multiscale spatiotemporal saliency value $SM_p^t$ at pixel $p$ at frame $t$ is given by:
\begin{equation}
SM_p^t = MCA\left( {\tilde S_{{\Omega _l}\left( p \right)}^t} \right),{\mkern 1mu} {\mkern 1mu} {\rm{with}}{\mkern 1mu} {\mkern 1mu} l \in \left\{ {1,2,...,L} \right\}{\mkern 1mu} ,
\label{equation:saliency_fusion}
\end{equation}
where $\Omega _l(\cdot)$ is a function that converts a pixel to the region at scale level $l$ where it belongs. We note that all operations are processed pixel-wisely. ${\tilde S_{{\Omega _l}\left( \cdot \right)}^t}$ measures multiscale saliency values at each pixel generated in the $l$-th scale of the segmentation model, which has $L$ scale levels, at frame $t$. $MCA(\cdot)$  is the Multi-layer Cellular Automata (MCA) integration\cite{Quin-CVPR2015}, which exploits intrinsic relevance of similar regions through interactions with neighbors. The MCA algorithm is described in Table \ref{tab:MCA}. Figure \ref{img:MCA} depicts examples of the MCA algorithm.



\section{Experimental settings}\label{section:experimental_setup}
\subsection{Datasets}\label{section:dataset}
We used 4 datasets for all experiments: Weizmann human action dataset\cite{Blank-ICCV2005}, MCL2014 dataset\cite{Lee-ICIP2014}, SegTrack2 dataset\cite{Li-ICCV2013}, and DAVIS dataset\cite{Perazzi-CVPR2016}.

\textbf{The Weizmann dataset}\cite{Blank-ICCV2005} contains 93 video sequences of nine people performing ten natural actions such as running, walking, jacking, and waving, etc. This dataset is created for human action recognition and has simple static background, thus it is easy to distinguish objects from scenes. Each sequence in the dataset has the spatial resolution of $180\times144$ and consists of about 60 frames with the ground-truth foreground mask for every frame.

\textbf{The MCL2014 dataset}\cite{Lee-ICIP2014} contains 8 video sequences with various background such as streets, roads, and halls, etc. In this dataset, multiple objects such as crowds move with different directions and speed. Each sequence in the dataset has the spatial resolution of $480\times270$ and consists of around 800 frames. The binary ground-truth maps are manually obtained for every 8 frames. We remark that another version of the MCL dataset was published and it is called MCL2015 dataset\cite{Hansang-TIP2015} which contains 9 video sequences. Due to the advantage of having double frames per video, the MCL2014 dataset (800 frames per video) was employed instead of the MCL2015 dataset (400 frames per video).

\textbf{The SegTrack2 dataset}\cite{Li-ICCV2013}, which is extended from the SegTrack dataset\cite{Tsai-IJCV2012}, contains 14 challenging video sequences and is originally designed for video object segmentation. A half of videos in this dataset have multiple salient objects. The dataset is designed to be challenging in that it has background-foreground color similarity, fast motion, and complex shape deformation. Dynamics in this dataset is caused by moving cameras, which track objects in scenes moving from far to near, from the borders to the center or vice versa. Each sequence in the dataset has the spatial resolution of $352\times288$ and consists of about 75 frames with the binary ground-truth mask.

\textbf{The DAVIS dataset}\cite{Perazzi-CVPR2016} consists of 50 high quality $854 \times 480$ spatial resolution and Full HD 1080p video sequences with about 70 frames per video, each of which has one single salient object or two spatially connected objects either with low contrast or overlapping with image boundary. It is a challenging dataset because of frequent occurrences of occlusions, motion blur and appearance changes. 
In this work, we used only $854 \times 480$ resolution video sequences.

We note that these datasets have different characteristics such as simple background for the Weizmann dataset, multiple moving objects for the MCL2014 dataset, diverse complex dynamic scenes for the SegTrack2 and the DAVIS datasets. All the datasets contain manually annotated pixel-wise ground-truth.

\subsection{Evaluation metrics}\label{metrics}

The \textbf{Precision-Recall}, and \textbf{F-measure} metrics are used to evaluate performance of the object location detection at a binary threshold. The precision value corresponds to the ratio of salient pixels that are correctly assigned to all the pixels of extracted regions. The recall value is defined as the percentage of detected salient pixels in relation to the number of salient pixels in the ground-truth. Given a ground-truth $GT$ and the binarized map $BM$ for a saliency map, we have:
\begin{equation}
Precision = \frac{{\left| {BM \cap GT} \right|}}{{\left| {BM} \right|}}{\mkern 1mu} {\mkern 1mu} {\mkern 1mu} {\mkern 1mu} {\rm{and}}{\mkern 1mu} {\mkern 1mu} {\mkern 1mu} {\mkern 1mu} Recall = \frac{{\left| {BM \cap GT} \right|}}{{\left| {GT} \right|}}.
\end{equation}

The F-measure\cite{Achanta-CVPR2009} is the overall performance measure computed by the weighted harmonic of precision and recall: 
\begin{equation}
{F_\beta } = \frac{{\left( {1 + {\beta ^2}} \right)Precision \times Recall}}{{{\beta ^2} \times Precision + Recall}}.
\end{equation}
Similarly to \cite{Achanta-CVPR2009}, we chose $\beta^2=0.3$ to weight precision more than recall. $F_\beta$ reflects the overall prediction accuracy. 

There are different ways of binarizing a saliency map to compute F-measure. We used the \textbf{F-Adap}, an adaptive threshold for generating a binary saliency map. As suggestion by \cite{Jia-ICCV2013}
, we employed an adaptive threshold for each image, which is determined as the sum of the average value and the standard deviation of the entire given saliency image: $\theta=\mu+\sigma$ where $\mu$ and $\sigma$ are the mean value and the standard deviation of the given saliency map, respectively. We then computed the average F-measure scores over the frames. We also used the \textbf{F-Max}, which describes the maximum F-measure score for different thresholds from 0 to 255, as suggested by \cite{Borji-TIP2015}. 

We note that for a threshold, we binarize the saliency map to compute Precision and Recall at each frame in a video and then take the average over the video. After that, the mean of the averages over videos in a dataset is computed. F-measure is computed from the final Precision and Recall.

Overlap-based evaluation measures mentioned above do not consider the true negative saliency assignments, i.e., the pixels correctly marked as non-salient\cite{Borji-TIP2013}. For a more comprehensive comparison, we thus used the \textbf{Mean Absolute Error (MAE)} to compute the average absolute per-pixel difference between a saliency map $SM$ and its corresponding ground-truth $GT$:
\begin{equation}
MAE = \frac{1}{{W \times H}}\sum\limits_{x = 1}^W {\sum\limits_{y = 1}^H {\left\| {SM\left( {x,y} \right) - GT\left( {x,y} \right)} \right\|} },
\end{equation}
where $W$ and $H$ are the width and the height of the maps, respectively. We note that MAE is also computed from mean average value of the dataset in the same way with F-measure.


\section{Comparison with state-of-the-art methods}\label{section:comparison}

We compared the performance of our method with several state-of-the-art models, namely, LC\cite{Zhai-MM2006}, LD\cite{Liu-PAMI2011}, RWRV\cite{Hansang-TIP2015}, SAG\cite{Wang-CVPR2015}, SEG\cite{Rahtu-ECCV2010}, STS\cite{Zhou-CVPR2014}, 
BMS\cite{Zhang-ICCV2013}, BSCA\cite{Quin-CVPR2015}, 
MBS\cite{Zhang-ICCV2015}, MC\cite{Jiang-ICCV2013}, 
MST\cite{Tu-CVPR2016}, 
and WSC\cite{Nianyi-CVPR2015}, which are classified in Table \ref{tab:compared_method}. We compared our method with not only video saliency models but also recent saliency detection methods for still images, which show good performance on image datasets. We frame-wisely applied the methods developed for the still image to videos. We remark that we run original codes provided by the authors with recommended parameter settings for obtaining results.

\begin{table}[t]
\centering
\caption{Compared state-of-the-art methods and classification.}
\label{tab:compared_method}
\begin{tabular}{|l|l|l|}
\hline
\textbf{target} & \multicolumn{1}{c|}{\textbf{image}} & \multicolumn{1}{c|}{\textbf{video}} \\ \hline
\textbf{method} & \begin{tabular}[c]{@{}l@{}}BL\cite{Tong-CVPR2015}, BSCA\cite{Quin-CVPR2015}, MBS\cite{Zhang-ICCV2015},\\ MC\cite{Jiang-ICCV2013}, MST\cite{Tu-CVPR2016}, WSC\cite{Nianyi-CVPR2015}\end{tabular} & \begin{tabular}[c]{@{}l@{}}LC\cite{Zhai-MM2006}, LD\cite{Liu-PAMI2011}, RWRV\cite{Hansang-TIP2015},\\ SAG\cite{Wang-CVPR2015}, SEG\cite{Rahtu-ECCV2010}, STS\cite{Zhou-CVPR2014}\end{tabular} \\ \hline
\end{tabular}
\end{table}
 
\begin{figure}[t]
    \centering
    \includegraphics[width=1\textwidth]{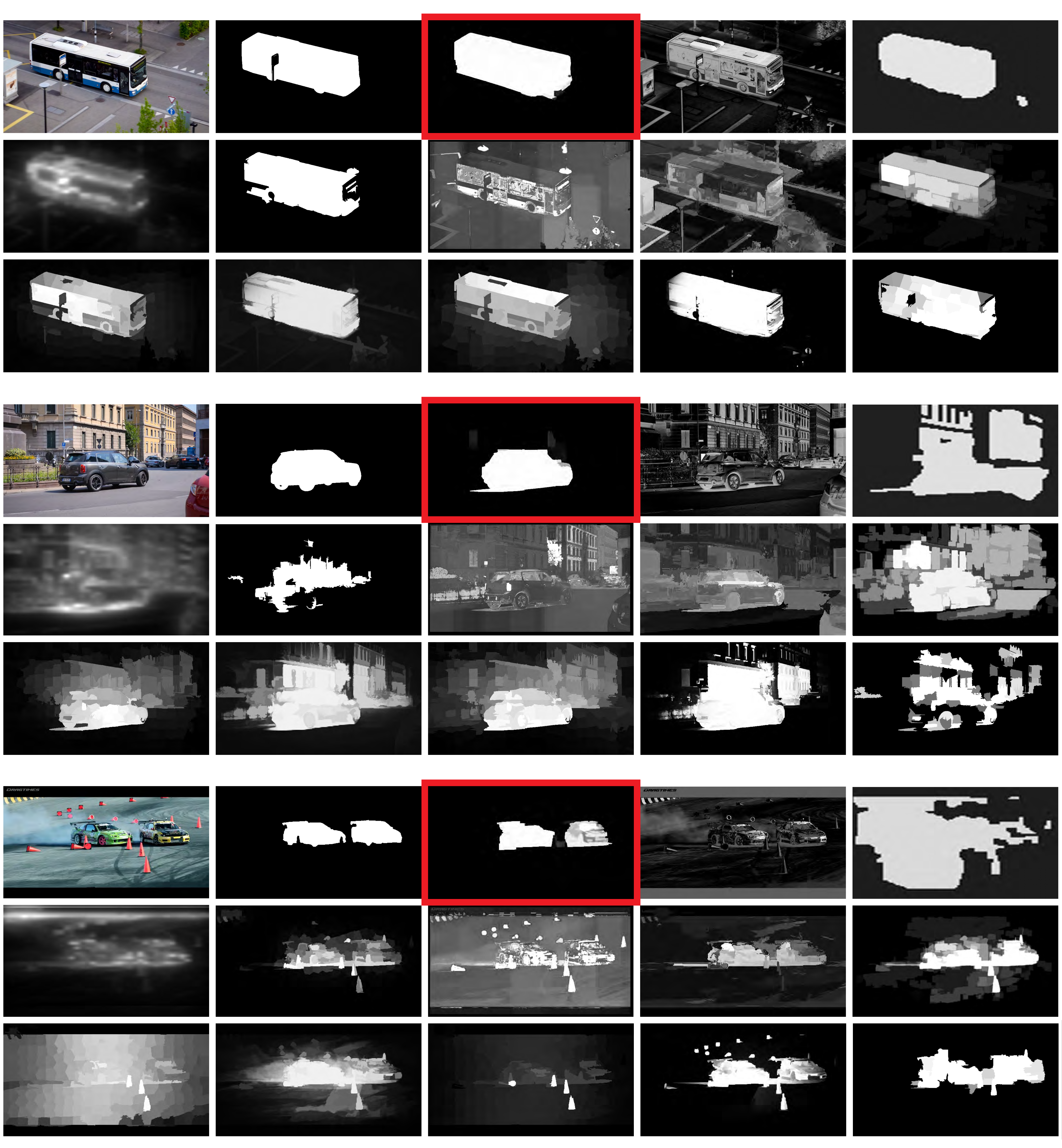}
    
    \caption{Visual comparison of our method to the state-of-the-art methods. From top-left to bottom-right, original image and ground-truth are followed by outputs obtained using our method, LC\protect\cite{Zhai-MM2006}, LD\protect\cite{Liu-PAMI2011}, RWRV\protect\cite{Hansang-TIP2015}, SAG\protect\cite{Wang-CVPR2015}, SEG\protect\cite{Rahtu-ECCV2010}, STS\protect\cite{Zhou-CVPR2014}, 
BMS\protect\cite{Zhang-ICCV2013}, BSCA\protect\cite{Quin-CVPR2015}, 
MBS\protect\cite{Zhang-ICCV2015}, MC\protect\cite{Jiang-ICCV2013}, 
MST\protect\cite{Tu-CVPR2016}, 
and WSC\protect\cite{Nianyi-CVPR2015}. Our method surrounded with red rectangles achieves the best results.}
\label{img:visual_comparison_1}
\end{figure}

\begin{figure}[t]
    \centering
    \includegraphics[width=1\textwidth]{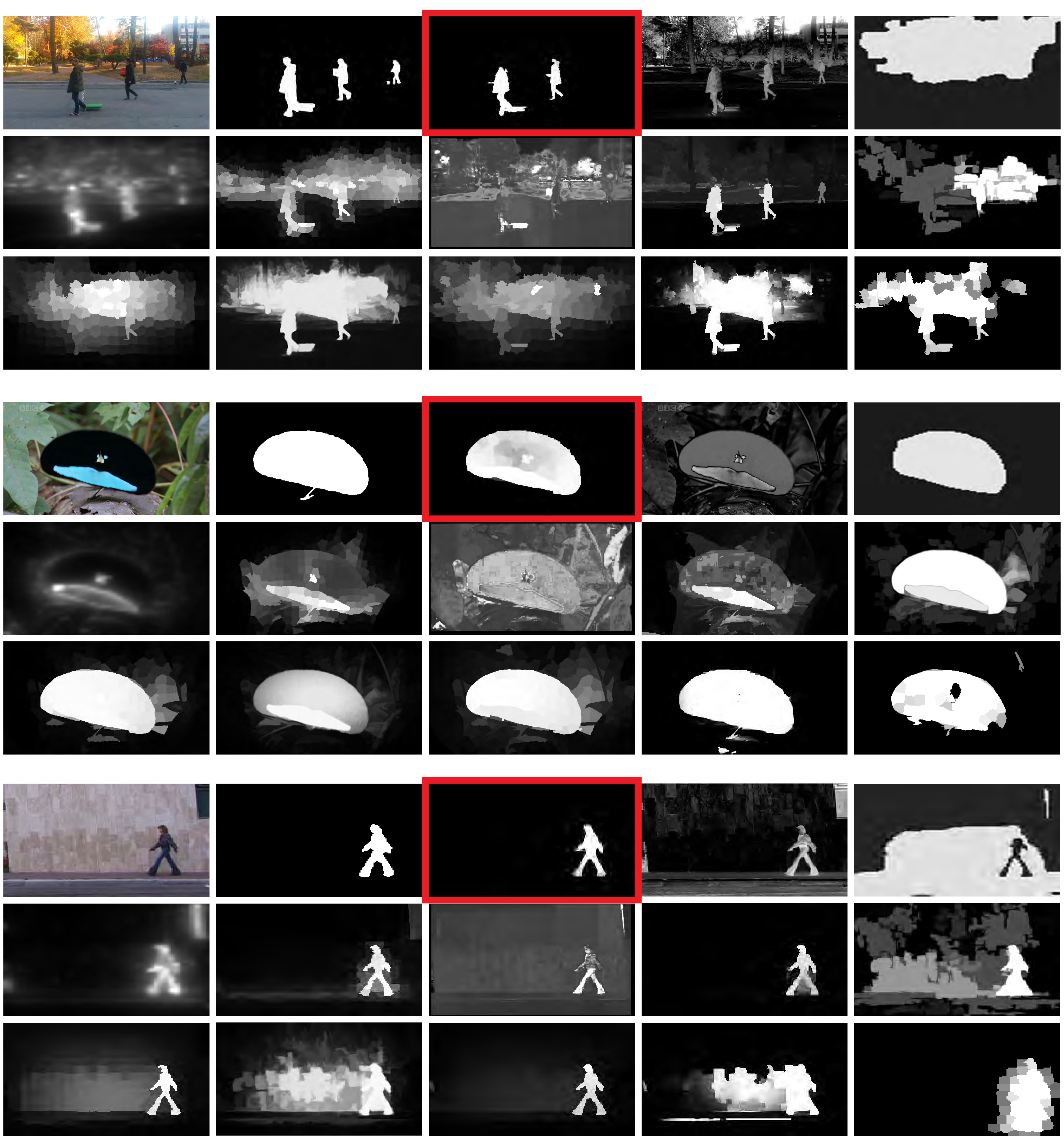}
    
    \caption{Visual comparison of our method to the state-of-the-art methods. From top-left to bottom-right, original image and ground-truth are followed by outputs obtained using our method, LC\protect\cite{Zhai-MM2006}, LD\protect\cite{Liu-PAMI2011}, RWRV\protect\cite{Hansang-TIP2015}, SAG\protect\cite{Wang-CVPR2015}, SEG\protect\cite{Rahtu-ECCV2010}, STS\protect\cite{Zhou-CVPR2014}, 
BMS\protect\cite{Zhang-ICCV2013}, BSCA\protect\cite{Quin-CVPR2015}, 
MBS\protect\cite{Zhang-ICCV2015}, MC\protect\cite{Jiang-ICCV2013}, 
MST\protect\cite{Tu-CVPR2016}, 
and WSC\protect\cite{Nianyi-CVPR2015}. Our method surrounded with red rectangles achieves the best results.}
\label{img:visual_comparison_2}
\end{figure}

Some examples for visual comparison of the methods are shown in Fig. \ref{img:visual_comparison_1} and Fig. \ref{img:visual_comparison_2}, suggesting that our method produces the best results on the datasets. Our method can handle complex foreground and background with different details, giving accurate and uniform saliency assignment. Our method also achieves the state-of-the-art substantially on the datasets on all evaluation criteria.

\begin{figure*}[t]
\footnotesize
    \centering
    \begin{tabular}{cc}
    \begin{tabular}[c]{@{}c@{}}\includegraphics[width=0.45\textwidth]{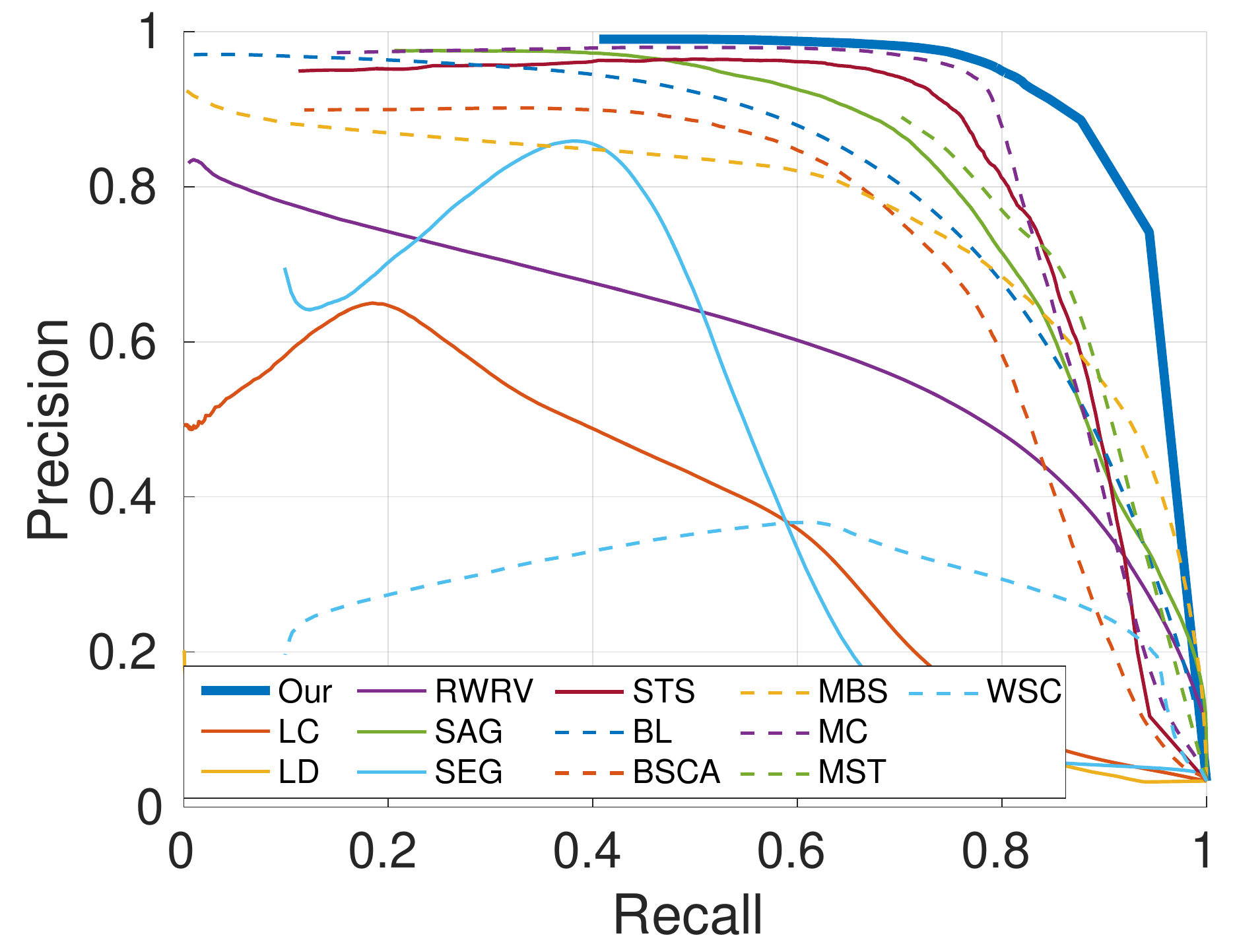} \\ (a)  Weizmann Dataset \end{tabular} & 
    \begin{tabular}[c]{@{}c@{}}\includegraphics[width=0.45\textwidth]{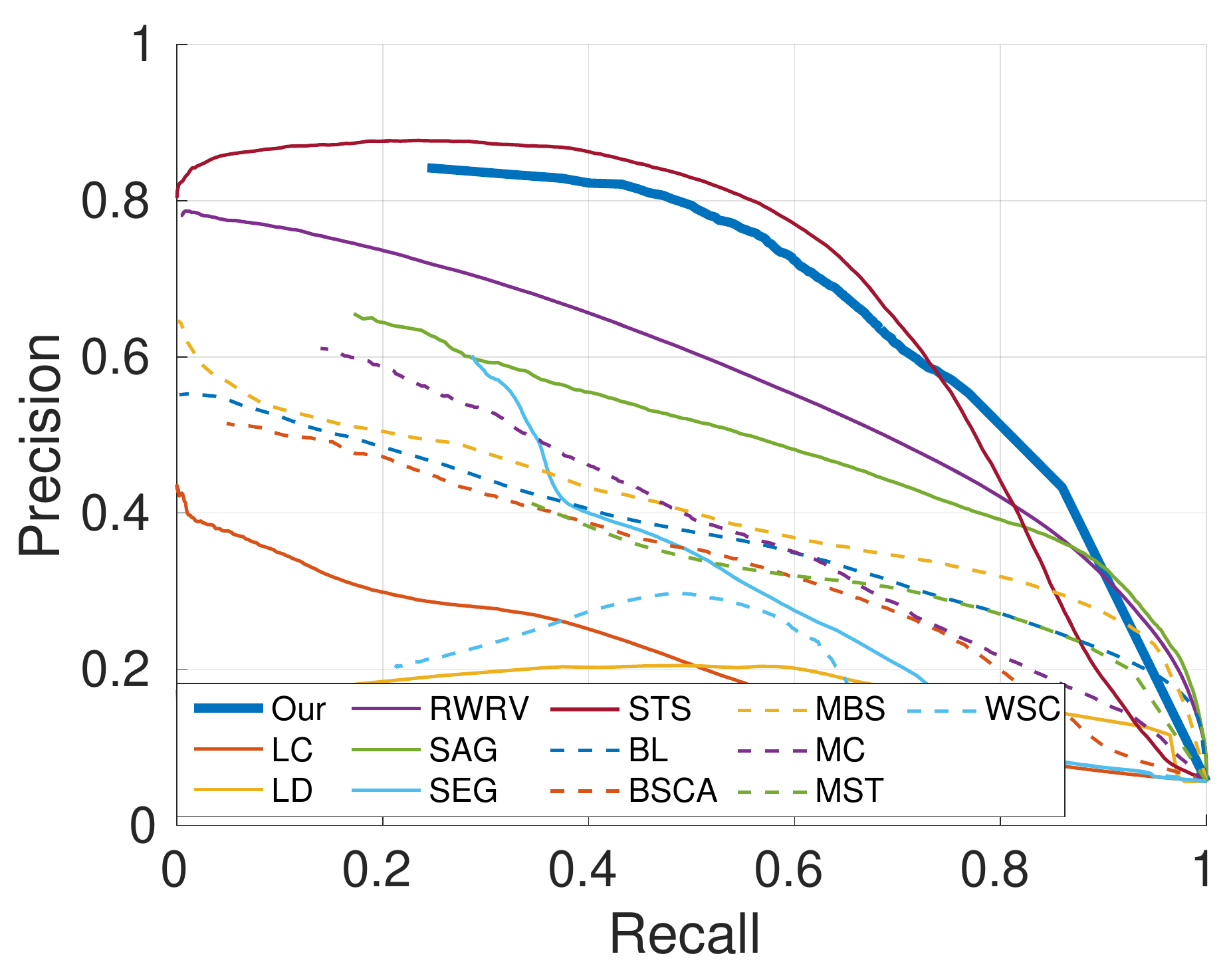} \\ (b)  MCL2014 Dataset \end{tabular} \cr
    \begin{tabular}[c]{@{}c@{}}\includegraphics[width=0.45\textwidth]{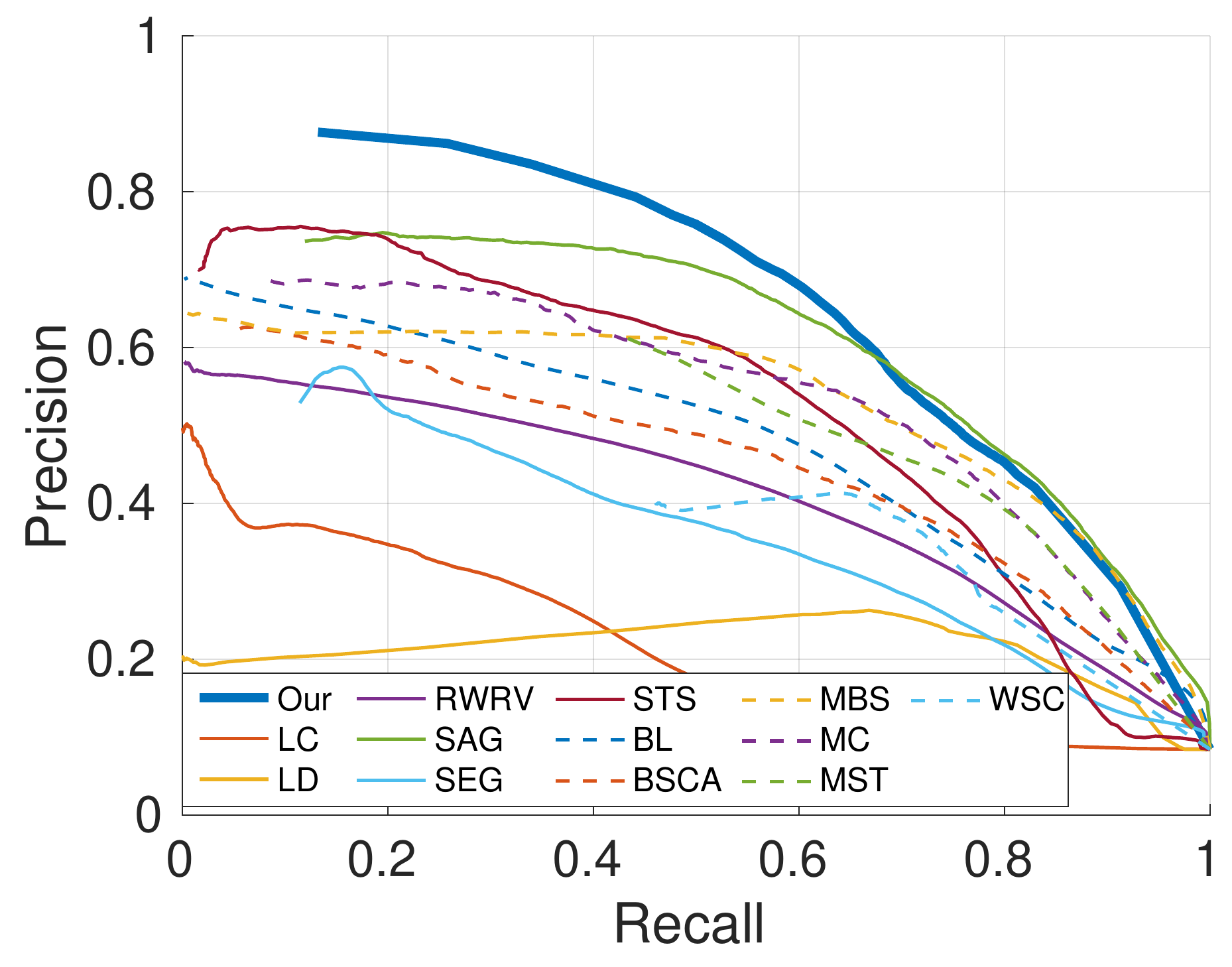} \\ (c)  SegTrack2 Dataset \end{tabular} & 
    \begin{tabular}[c]{@{}c@{}}\includegraphics[width=0.45\textwidth]{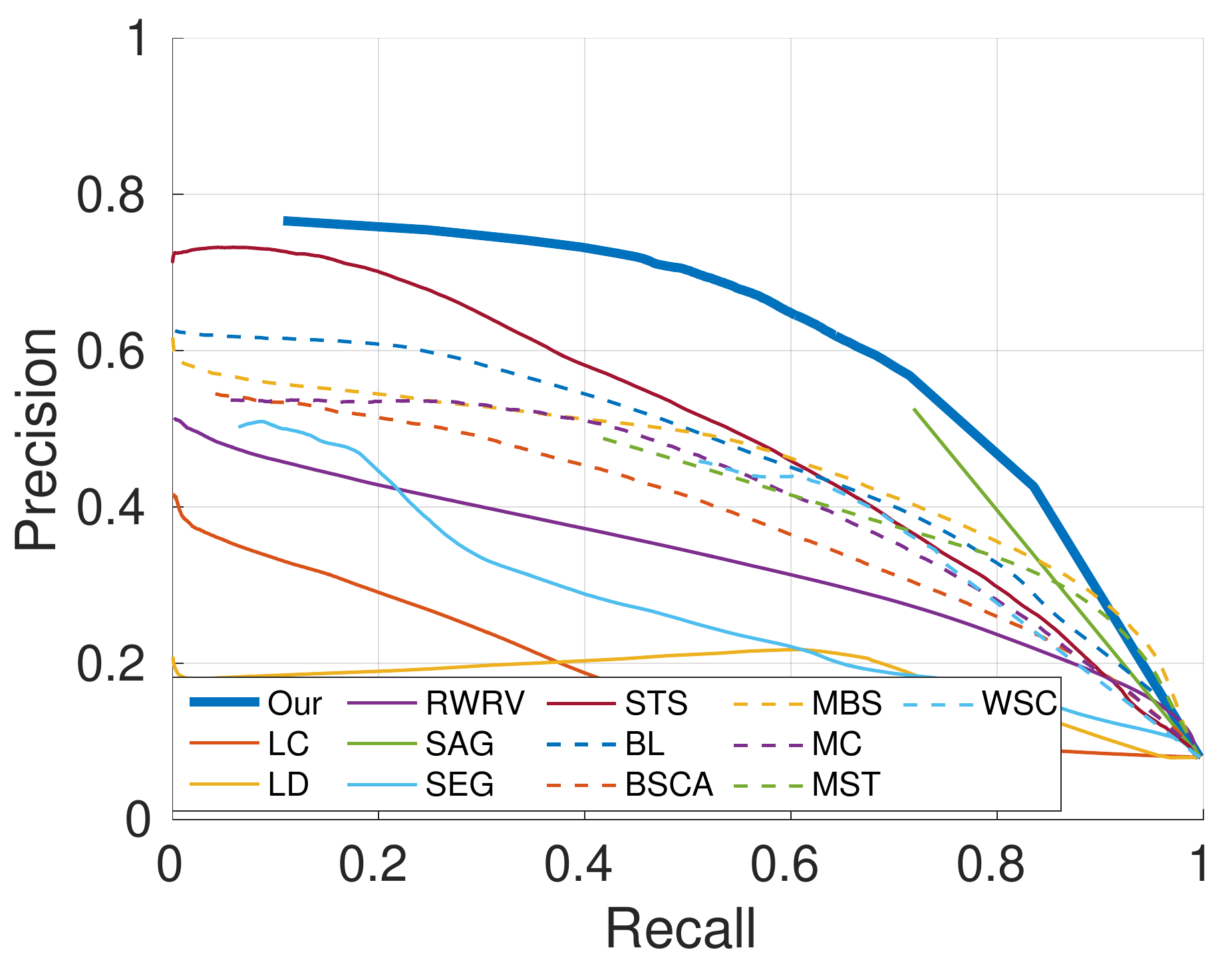} \\ (d)  DAVIS Dataset \end{tabular} \cr
    \end{tabular}
    \caption{Quantitative comparison with state-of-the-art methods, using Precision-Recall Curves at different fixed thresholds. Our method is marked in \textbf{bold curve}. Note that the state-of-the-art methods are divided into two groups only for the clear presentation.}
\label{fig:PRC}
\end{figure*}

\subsection{Precision-Recall Curve}

We binarized each saliency map into a binary mask using a binary threshold $\theta$ ($\theta$ is changed from $0$ to $255$). With each $\theta$, the binary mask is checked against the ground-truth to evaluate the accuracy of the salient object detection to compute the Precision-Recall Curve (PRC) (c.f. Fig. \ref{fig:PRC}). The PRC  is used to evaluate the performance of the object location detection because it captures behaviors of both precision and recall under varying thresholds. Therefore, the PRC provides a reliable comparison of how well various saliency maps can highlight salient regions in images. 

Figure \ref{fig:PRC} shows that the proposed method consistently produces saliency maps closer to the ground-truth than the others. Our method achieves the highest precision in most of entire recall ranges on almost datasets. We observe that our method is the best method on the Weizmann dataset, the SegTrack2 dataset, and the DAVIS dataset while on the MCL2014 dataset, our method is the second to the best (lower than STS\cite{Zhou-CVPR2014}). Therefore, the proposed method significantly outperforms the state-of-the-art methods across the public datasets. Especially in the most challenging datasets (e.g. SegTrack2 dataset and DAVIS dataset), the performance gains of our method over all the other methods are more noticeable. Our method is suitable for dealing with dynamic scenes and multiple moving objects in videos.

From Figure \ref{fig:PRC}, at the first end of PRCs at maximum recall, all salient pixels are retained as positives, i.e., considered to be foreground, so all the methods have the same precision and recall values. At the other end of PRCs, our method is one of the methods whose minimum recall values are highest on datasets, meaning that our method generates saliency maps containing more salient pixels with the maximum saliency values at 1. Therefore, our method has more practical advantages than other methods. For example, when we cannot decide the best binary threshold to extract salient objects from a saliency map, an adaptive threshold of the saliency map can be used, but the accuracy of object extraction is still ensured (c.f. Section \ref{section:adaptive_threshold}).

\subsection{F-measure}\label{section:adaptive_threshold}


We evaluated F-Adap and F-Max to compare the proposed method with the state-of-the-art methods (c.f. Table \ref{tab:comparison_F}). Table \ref{tab:comparison_F} shows that our method achieves the best performance on all the datasets. Our method achieves the highest scores on the Weizmann dataset, the SegTrack2 dataset, and the DAVIS dataset on both the metrics. On the most challenging dataset, the DAVIS dataset according to \cite{Perazzi-CVPR2016}, the proposed method outperforms the second best method by a large margin on all the metrics. Our method achieves 0.627 in the F-Adap and 0.645 in the F-Max while the second best method (SAG\cite{Wang-CVPR2015}) achieves 0.494 and 0.548, respectively. On the MCL2014 dataset, the proposed method achieves 0.644 in the F-Adap while other methods do lower than 0.5. Our method is the second best in the F-Max and slightly lower than the best method (STS\cite{Zhou-CVPR2014}) (0.702 vs. 0.728).

\begin{table*}[t]
\centering
\caption{Quantitative comparison with state-of-the-art methods on three datasets, using F-measure (F-Adap and F-Max) (higher is better). The best and the second best results are shown in \textcolor[rgb]{0,0,1}{blue} and \textcolor[rgb]{0,0.7,0}{green}, respectively. Our method is marked in \textbf{bold}.}
\label{tab:comparison_F}
\resizebox{1\linewidth}{!}{%
\begin{tabular}{lcclcclcclcc}
\hline \hline
\textbf{Dataset} & 
\multicolumn{2}{c}{\textbf{Weizmann\cite{Blank-ICCV2005}}} & \multicolumn{1}{c}{\textbf{}} &  
\multicolumn{2}{c}{\textbf{MCL2014\cite{Lee-ICIP2014}}}    & \multicolumn{1}{c}{\textbf{}} & 
\multicolumn{2}{c}{\textbf{SegTrack2\cite{Li-ICCV2013}}}   & \multicolumn{1}{c}{\textbf{}} & 
\multicolumn{2}{c}{\textbf{DAVIS\cite{Perazzi-CVPR2016}}}  \\ \cline{2-3} \cline{5-6} \cline{8-9} \cline{11-12}

\textbf{Metric} & 
\multicolumn{1}{c}{\textbf{F-Adap}} & \multicolumn{1}{c}{\textbf{F-Max}} & \multicolumn{1}{c}{\textbf{}} & 
\multicolumn{1}{c}{\textbf{F-Adap}} & \multicolumn{1}{c}{\textbf{F-Max}} & \multicolumn{1}{c}{\textbf{}} & 
\multicolumn{1}{c}{\textbf{F-Adap}} & \multicolumn{1}{c}{\textbf{F-Max}} & \multicolumn{1}{c}{\textbf{}} & 
\multicolumn{1}{c}{\textbf{F-Adap}} & \multicolumn{1}{c}{\textbf{F-Max}} \\ \hline

\textbf{Our}                  & \textcolor[rgb]{0,0,1}{0.909} & \textcolor[rgb]{0,0,1}{0.912} & & \textcolor[rgb]{0,0,1}{0.644} & \textcolor[rgb]{0,0.7,0}{0.702} & & \textcolor[rgb]{0,0,1}{0.510} & \textcolor[rgb]{0,0,1}{0.677} & & \textcolor[rgb]{0,0,1}{0.627} & \textcolor[rgb]{0,0,1}{0.645} \\
LC\cite{Zhai-MM2006}          & 0.243 & 0.467 & & 0.191 & 0.285 & & 0.244 & 0.306 & & 0.201 & 0.265 \\
LD\cite{Liu-PAMI2011}         & 0.098 & 0.112 & & 0.228 & 0.239 & & 0.286 & 0.305 & & 0.252 & 0.256 \\
RWRV\cite{Hansang-TIP2015}    & 0.388 & 0.604 & & 0.390 & 0.579 & & 0.355 & 0.463 & & 0.318 & 0.378 \\
SAG\cite{Wang-CVPR2015}       & 0.560 & 0.828 & & 0.400 & 0.515 & & \textcolor[rgb]{0,0.7,0}{0.504} & \textcolor[rgb]{0,0.7,0}{0.646} & & \textcolor[rgb]{0,0.7,0}{0.494} & \textcolor[rgb]{0,0.7,0}{0.548} \\
SEG\cite{Rahtu-ECCV2010}      & 0.616 & 0.683 & & 0.372 & 0.480 & & 0.388 & 0.418 & & 0.305 & 0.348 \\
STS\cite{Zhou-CVPR2014}       & \textcolor[rgb]{0,0.7,0}{0.756} & 0.872 & & \textcolor[rgb]{0,0.7,0}{0.500} & \textcolor[rgb]{0,0,1}{0.728} & & 0.471 & 0.583 & & 0.379 & 0.527 \\ 

BL\cite{Tong-CVPR2015}        & 0.255 & 0.794 & & 0.326 & 0.405 & & 0.370 & 0.520 & & 0.387 & 0.505 \\
BSCA\cite{Quin-CVPR2015}      & 0.464 & 0.774 & & 0.327 & 0.392 & & 0.357 & 0.492 & & 0.377 & 0.441 \\
MBS\cite{Zhang-ICCV2015}      & 0.463 & 0.761 & & 0.372 & 0.427 & & 0.418 & 0.582 & & 0.393 & 0.498 \\
MC\cite{Jiang-ICCV2013}       & 0.747 & \textcolor[rgb]{0,0.7,0}{0.899} & & 0.358 & 0.454 & & 0.420 & 0.566 & & 0.422 & 0.483 \\
MST\cite{Tu-CVPR2016}         & 0.569 & 0.838 & & 0.352 & 0.395 & & 0.475 & 0.559 & & 0.395 & 0.469 \\
WSC\cite{Nianyi-CVPR2015}     & 0.265 & 0.405 & & 0.284 & 0.327 & & 0.370 & 0.450 & & 0.408 & 0.473 \\

\hline \hline
\end{tabular}
}
\end{table*}

\subsection{Mean Absolute Error}

To further demonstrate the effectiveness of the proposed method, we provide the comparison of our method with the state-of-the-art methods using the MAE metric. Table \ref{tab:comparison_MAE} shows results of our method and the other methods on the four benchmark datasets. Our method outperforms the other methods on almost all datasets. The MAE of our method is lower than the others, which suggests our method not only highlights the overall salient objects but also preserves the detail better. It can be seen from Table \ref{tab:comparison_MAE} that our method shows the lowest MAE on the Weizmann dataset, the MCL2014 dataset, and the DAVIS dataset. Our method achieves 0.009, 0.051, and 0.077 in the MEA on the Weizmann dataset, the MCL2014 dataset, and the DAVIS dataset while the second best methods achieve 0.012 for MBS\cite{Zhang-ICCV2015}, 0.107 for STS\cite{Zhou-CVPR2014}, and 0.103 for SAG\cite{Wang-CVPR2015}, respectively. Our method outperforms the second best methods by a large margin on these datasets. For the SegTrack2 dataset, our method is the second best method at 0.125 while the second best method, SAG\cite{Wang-CVPR2015}, is at 0.106.

\begin{table}[t]
\centering
\caption{Quantitative comparison with state-of-the-art methods on three datasets, using Mean Absolute Errors (MAE) (smaller is better). The best and the second best results are shown in \textcolor[rgb]{0,0,1}{blue} and \textcolor[rgb]{0,0.7,0}{green}, respectively. Our method is marked in \textbf{bold}.}
\label{tab:comparison_MAE}
\begin{tabular}{lcccc}
\hline \hline

\textbf{Dataset} & \textbf{Weizmann\cite{Blank-ICCV2005}} & \textbf{MCL2014\cite{Lee-ICIP2014}} & \textbf{SegTrack2\cite{Li-ICCV2013}} & \textbf{DAVIS\cite{Perazzi-CVPR2016}}  \\ \hline

\textbf{Our} & \textcolor[rgb]{0,0,1}{0.009} & \textcolor[rgb]{0,0,1}{0.051} & \textcolor[rgb]{0,0.7,0}{0.125} &  \textcolor[rgb]{0,0,1}{0.077} \\

LC\cite{Zhai-MM2006}          & 0.135 & 0.195 & 0.173 & 0.191 \\
LD\cite{Liu-PAMI2011}         & 0.376 & 0.290 & 0.281 & 0.302 \\
RWRV\cite{Hansang-TIP2015}    & 0.093 & 0.155 & 0.185 & 0.200 \\
SAG\cite{Wang-CVPR2015}       & 0.045 & 0.131 & \textcolor[rgb]{0,0,1}{0.106} & \textcolor[rgb]{0,0.7,0}{0.103} \\
SEG\cite{Rahtu-ECCV2010}      & 0.246 & 0.288 & 0.321 & 0.323 \\
STS\cite{Zhou-CVPR2014}       & 0.024 & \textcolor[rgb]{0,0.7,0}{0.107} & 0.147 & 0.183 \\ 

BL\cite{Tong-CVPR2015}        & 0.300 & 0.240 & 0.286 & 0.223 \\
BSCA\cite{Quin-CVPR2015}      & 0.150 & 0.201 & 0.209 & 0.197 \\
MBS\cite{Zhang-ICCV2015}      & \textcolor[rgb]{0,0.7,0}{0.012} & 0.215 & 0.190 & 0.225 \\
MC\cite{Jiang-ICCV2013}       & 0.089 & 0.196 & 0.196 & 0.182 \\
MST\cite{Tu-CVPR2016}         & 0.043 & 0.181 & 0.142 & 0.175 \\
WSC\cite{Nianyi-CVPR2015}     & 0.107 & 0.159 & 0.136 & 0.120 \\

\hline \hline
\end{tabular}
\end{table}


\section{Evaluation of the proposed method}\label{section:validation}

In this part, we analyzed and discussed each component of the proposed method to evaluate actual contribution of each component. For the evaluation of the proposed method, we used F-Adap, F-Max, and MAE.

\subsection{Multiple level processing evaluation}

\begin{table}[th]
\centering
\caption{Multiple level processing evaluation, using F-measure scores (higher is better) and Mean Absolute Error (smaller is better). The best results are shown in \textcolor[rgb]{0,0,1}{blue}.}
\label{tab:multiscale_experiment}
\begin{tabular}{l|lllllll}
\hline \hline
\textbf{Dataset}   & \textbf{Method} &  & \textbf{F-Adap $\Uparrow$} &  & \textbf{F-Max $\Uparrow$} &  & \textbf{MAE $\Downarrow$} \\ \hline
\textbf{Weizmann}  & \begin{tabular}[c]{@{}l@{}}Multi-level\\ $1^{st}$ Level\\ $2^{nd}$ Level\\ $3^{rd}$ Level\end{tabular} & & 
\begin{tabular}[c]{@{}l@{}}\textcolor[rgb]{0,0,1}{0.909}\\ 0.773\\ 0.748\\ 0.727\end{tabular} & & 
\begin{tabular}[c]{@{}l@{}}\textcolor[rgb]{0,0,1}{0.912}\\ 0.904\\ 0.902\\ 0.897\end{tabular} & & 
\begin{tabular}[c]{@{}l@{}}\textcolor[rgb]{0,0,1}{0.009}\\ 0.040\\ 0.039\\ 0.040\end{tabular} \\ \hline
\textbf{MCL2014}   & \begin{tabular}[c]{@{}l@{}}Multi-level\\ $1^{st}$ Level\\ $2^{nd}$ Level\\ $3^{rd}$ Level\end{tabular} & & 
\begin{tabular}[c]{@{}l@{}}\textcolor[rgb]{0,0,1}{0.644}\\ 0.557\\ 0.565\\ 0.565\end{tabular} & & 
\begin{tabular}[c]{@{}l@{}}\textcolor[rgb]{0,0,1}{0.702}\\ 0.632\\ 0.688\\ 0.702\end{tabular} & & 
\begin{tabular}[c]{@{}l@{}}\textcolor[rgb]{0,0,1}{0.051}\\ 0.114\\ 0.108\\ 0.106\end{tabular} \\ \hline
\textbf{SegTrack2} & \begin{tabular}[c]{@{}l@{}}Multi-level\\ $1^{st}$ Level\\ $2^{nd}$ Level\\ $3^{rd}$ Level\end{tabular} & & 
\begin{tabular}[c]{@{}l@{}}\textcolor[rgb]{0,0,1}{0.510}\\ 0.463\\ 0.460\\ 0.465\end{tabular} & & 
\begin{tabular}[c]{@{}l@{}}\textcolor[rgb]{0,0,1}{0.677}\\ 0.667\\ 0.669\\ 0.667\end{tabular} & & 
\begin{tabular}[c]{@{}l@{}}\textcolor[rgb]{0,0,1}{0.125}\\ 0.175\\ 0.175\\ 0.175\end{tabular} \\ \hline
\textbf{DAVIS}     & \begin{tabular}[c]{@{}l@{}}Multi-level\\ $1^{st}$ Level\\ $2^{nd}$ Level\\ $3^{rd}$ Level\end{tabular} & & 
\begin{tabular}[c]{@{}l@{}}\textcolor[rgb]{0,0,1}{0.627}\\ 0.554\\ 0.556\\ 0.550\end{tabular} & & 
\begin{tabular}[c]{@{}l@{}}\textcolor[rgb]{0,0,1}{0.645}\\ 0.640\\ 0.642\\ 0.638\end{tabular} & & 
\begin{tabular}[c]{@{}l@{}}\textcolor[rgb]{0,0,1}{0.077}\\ 0.142\\ 0.143\\ 0.146\end{tabular} \\ \hline
\textbf{Average}   & \begin{tabular}[c]{@{}l@{}}Multi-level\\ $1^{st}$ Level\\ $2^{nd}$ Level\\ $3^{rd}$ Level\end{tabular} &  &
\begin{tabular}[c]{@{}l@{}}\textcolor[rgb]{0,0,1}{0.673}\\ 0.587\\ 0.582\\ 0.577\end{tabular} & & 
\begin{tabular}[c]{@{}l@{}}\textcolor[rgb]{0,0,1}{0.734}\\ 0.711\\ 0.725\\ 0.726\end{tabular} & & 
\begin{tabular}[c]{@{}l@{}}\textcolor[rgb]{0,0,1}{0.066}\\ 0.118\\ 0.116\\ 0.117\end{tabular} \\ \hline \hline
\end{tabular}
\end{table}

To verify effectiveness of our introduced multiple scale analysis, we performed experiments to compare methods in cases: analyzing saliency cues in a multiscale segmentation model through combining saliency maps at 3 levels (denoted by \textbf{Multi-level}) and in single scales separately (denoted by \textbf{$1^{st}$ Level}, \textbf{$2^{nd}$ Level}, and \textbf{$3^{rd}$ Level}).

The results in Table \ref{tab:multiscale_experiment} show that using multiple scale analysis outperforms saliency computation at a single scale level on all datasets. Processing in multiple scales yields much improvement over the method with a single layer saliency computation. Therefore, utilizing information from multiple image layers makes our method gain benefit.

\subsection{Evaluation of combination of low-level feature map and middle-level feature map}\label{section:validation_combination}

\begin{table}[t]
\centering
\caption{Feature map combination evaluation, using F-measure scores (higher is better) and Mean Absolute Error (smaller is better). The best results are shown in \textcolor[rgb]{0,0,1}{blue}.}
\label{tab:comparison_experiment}
\resizebox{1\linewidth}{!}{%
\begin{tabular}{l|lllllll}
\hline \hline
\textbf{Dataset}   & \textbf{Method} &  & \textbf{F-Adap $\Uparrow$} &  & \textbf{F-Max $\Uparrow$} &  & \textbf{MAE $\Downarrow$} \\ \hline
\textbf{Weizmann}  & \begin{tabular}[c]{@{}l@{}}Feature combination\\ Low-level feature\\ Middle-level feature\end{tabular} &  & 
\begin{tabular}[c]{@{}l@{}}\textcolor[rgb]{0,0,1}{0.909}\\ 0.412\\ 0.740\end{tabular} &  & 
\begin{tabular}[c]{@{}l@{}}\textcolor[rgb]{0,0,1}{0.912}\\ 0.876\\ 0.894\end{tabular} &  & 
\begin{tabular}[c]{@{}l@{}}\textcolor[rgb]{0,0,1}{0.009}\\ 0.153\\ 0.030\end{tabular} \\ \hline
\textbf{MCL2014}   & \begin{tabular}[c]{@{}l@{}}Feature combination\\ Low-level feature\\ Middle-level feature\end{tabular} &  & 
\begin{tabular}[c]{@{}l@{}}\textcolor[rgb]{0,0,1}{0.644}\\ 0.496\\ 0.440\end{tabular} &  & 
\begin{tabular}[c]{@{}l@{}}\textcolor[rgb]{0,0,1}{0.702}\\ 0.588\\ 0.611\end{tabular} &  & 
\begin{tabular}[c]{@{}l@{}}\textcolor[rgb]{0,0,1}{0.051}\\ 0.121\\ 0.146\end{tabular} \\ \hline
\textbf{SegTrack2} & \begin{tabular}[c]{@{}l@{}}Feature combination\\ Low-level feature\\ Middle-level feature\end{tabular} &  & 
\begin{tabular}[c]{@{}l@{}}\textcolor[rgb]{0,0,1}{0.510}\\ 0.360\\ 0.359\end{tabular} &  & 
\begin{tabular}[c]{@{}l@{}}\textcolor[rgb]{0,0,1}{0.677}\\ 0.491\\ 0.549\end{tabular} &  & 
\begin{tabular}[c]{@{}l@{}}\textcolor[rgb]{0,0,1}{0.125}\\ 0.238\\ 0.197\end{tabular} \\ \hline
\textbf{DAVIS}     & \begin{tabular}[c]{@{}l@{}}Feature combination\\ Low-level feature\\ Middle-level feature\end{tabular} &  & 
\begin{tabular}[c]{@{}l@{}}\textcolor[rgb]{0,0,1}{0.627}\\ 0.526\\ 0.303\end{tabular} &  & 
\begin{tabular}[c]{@{}l@{}}\textcolor[rgb]{0,0,1}{0.645}\\ 0.560\\ 0.392\end{tabular} &  & 
\begin{tabular}[c]{@{}l@{}}\textcolor[rgb]{0,0,1}{0.077}\\ 0.119\\ 0.249\end{tabular} \\ \hline
\textbf{Average}   & \begin{tabular}[c]{@{}l@{}}Feature combination\\ Low-level feature\\ Middle-level feature\end{tabular} &  & 
\begin{tabular}[c]{@{}l@{}}\textcolor[rgb]{0,0,1}{0.673}\\ 0.449\\ 0.461\end{tabular} &  & 
\begin{tabular}[c]{@{}l@{}}\textcolor[rgb]{0,0,1}{0.734}\\ 0.629\\ 0.612\end{tabular} &  & 
\begin{tabular}[c]{@{}l@{}}\textcolor[rgb]{0,0,1}{0.066}\\ 0.158\\ 0.156\end{tabular} \\ \hline \hline
\end{tabular}
}
\end{table}

In order to verify the effectiveness of combining low-level and middle-level features to generate a saliency map, we conducted experiments to compare our method (denoted by \textbf{Feature combination}) with the methods using a single kind of feature map (denoted by \textbf{Low-level feature} (the integration weight $\alpha=1$) and \textbf{Middle-level feature} ($\alpha=0$)). Table \ref{tab:comparison_experiment} illustrates the results, showing that combining both two kinds of feature maps significantly outperforms using a single feature map separately on all datasets. Therefore, utilizing multiple features yields much improvement over the method using only a single feature.

\subsection{Temporal consistency evaluation}

\begin{table}[t]
\centering
\caption{Temporal consistency evaluation, using F-measure scores (higher is better) and Mean Absolute Error (smaller is better). The best results are shown in \textcolor[rgb]{0,0,1}{blue}.}
\label{tab:ATW_experiment}
\resizebox{\linewidth}{!}{%
\begin{tabular}{l|lllllll}
\hline \hline
\textbf{Dataset}   & \textbf{Method} &  & \textbf{F-Adap $\Uparrow$} &  & \textbf{F-Max $\Uparrow$} &  & \textbf{MAE $\Downarrow$} \\ \hline
\textbf{Weizmann}  & \begin{tabular}[c]{@{}l@{}}with Temporal consistency\\ w/o Temporal consistency\end{tabular} &  & 
\begin{tabular}[c]{@{}l@{}}\textcolor[rgb]{0,0,1}{0.909}\\ 0.899\end{tabular} &  & 
\begin{tabular}[c]{@{}l@{}}\textcolor[rgb]{0,0,1}{0.912}\\ 0.907\end{tabular} &  & 
\begin{tabular}[c]{@{}l@{}}\textcolor[rgb]{0,0,1}{0.009}\\ 0.010\end{tabular} \\ \hline
\textbf{MCL2014}   & \begin{tabular}[c]{@{}l@{}}with Temporal consistency\\ w/o Temporal consistency\end{tabular} &  & 
\begin{tabular}[c]{@{}l@{}}0.644\\ \textcolor[rgb]{0,0,1}{0.645}\end{tabular} &  & 
\begin{tabular}[c]{@{}l@{}}0.702\\ \textcolor[rgb]{0,0,1}{0.703}\end{tabular} &  & 
\begin{tabular}[c]{@{}l@{}}\textcolor[rgb]{0,0,1}{0.051}\\ 0.051\end{tabular} \\ \hline
\textbf{SegTrack2} & \begin{tabular}[c]{@{}l@{}}with Temporal consistency\\ w/o Temporal consistency\end{tabular} &  & 
\begin{tabular}[c]{@{}l@{}}0.510\\ \textcolor[rgb]{0,0,1}{0.515}\end{tabular} &  & 
\begin{tabular}[c]{@{}l@{}}0.677\\ \textcolor[rgb]{0,0,1}{0.680}\end{tabular} &  & 
\begin{tabular}[c]{@{}l@{}}0.125\\ \textcolor[rgb]{0,0,1}{0.122}\end{tabular} \\ \hline
\textbf{DAVIS}     & \begin{tabular}[c]{@{}l@{}}with Temporal consistency\\ w/o Temporal consistency\end{tabular} &  & 
\begin{tabular}[c]{@{}l@{}}\textcolor[rgb]{0,0,1}{0.627}\\ 0.624\end{tabular} &  & 
\begin{tabular}[c]{@{}l@{}}\textcolor[rgb]{0,0,1}{0.645}\\ 0.643\end{tabular} &  & 
\begin{tabular}[c]{@{}l@{}}\textcolor[rgb]{0,0,1}{0.077}\\ 0.079\end{tabular} \\ \hline
\textbf{Average}   & \begin{tabular}[c]{@{}l@{}}with Temporal consistency\\ w/o Temporal consistency\end{tabular} &  &
\begin{tabular}[c]{@{}l@{}}\textcolor[rgb]{0,0,1}{0.673}\\ 0.671\end{tabular} &  & 
\begin{tabular}[c]{@{}l@{}}\textcolor[rgb]{0,0,1}{0.734}\\ 0.733\end{tabular} &  & 
\begin{tabular}[c]{@{}l@{}}\textcolor[rgb]{0,0,1}{0.066}\\ 0.066\end{tabular} \\ \hline \hline
\end{tabular}%
}
\end{table}

We evaluated the effectiveness of introducing ATW by comparing our method (denoted by \textbf{with Temporal consistency}) with the method not incorporating temporal consistency (denoted by \textbf{w/o Temporal consistency}). Table \ref{tab:ATW_experiment} illustrates the results. We observe that the ATW introduction slightly improves results on the Weizmann dataset and the DAVIS dataset, while it slightly degrades results on the MCL2014 dataset and the SegTrack2 dataset. However, the averaging results still show that the ATW introduction improves the performance of the proposed method.

\begin{figure}[t]
    \centering
    \includegraphics[width=0.85\linewidth]{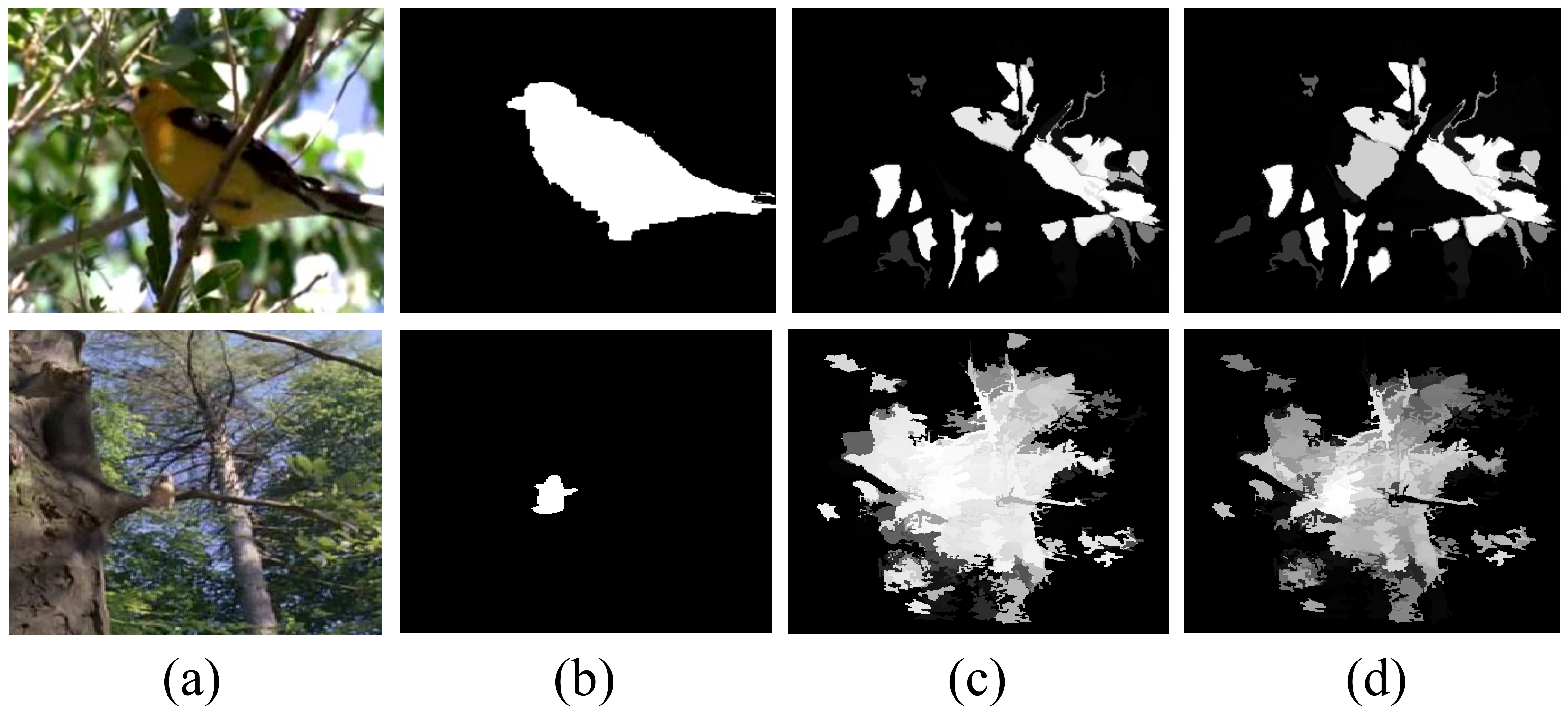}
    \caption{Failure cases of our method. (a) Input frames, (b) Ground-truth, (c) Results of our method, (d) Results of not incorporating temporal consistency.}
\label{fig:limitation}
\end{figure}

The limitation of ATW incorporation seems dependent on the accuracy of superpixel segmentation. Ineffectively propagating superpixels across time causes combining saliency values between incorrect regions, which leads to propagating errors in temporal consistency incorporation process. In Fig. \ref{fig:limitation}, we present some failure cases. In a video sequence having extremely difficult scenes with small, low-contrast salient objects (see examples in Fig. \ref{fig:limitation}), temporal superpixel methods cannot effectively segment the scene and cannot correctly track region over frames. This will result in poor quality in salient object detection. Firstly, spatial saliency entities fail in distinguishing foreground regions from the background when they are similar in appearance at almost frames. Secondly, errors can be propagated when combining temporal consistency from incorrect regions, resulting in worse spatiotemporal saliency maps than processing with a single frame.   



\section{Conclusion}\label{section:conclusion}

Differently from images, videos include dynamics, i.e., temporal position changes of entities in each frame and temporal changes of background. Such dynamics should be significantly considered when detecting salient objects in videos. To capture the dynamics, regions corresponding to each entity in the video frame are more suitable than pixels. Motivated by this, we present a region-based multiscale spatiotemporal saliency detection method for videos. In our method, we first segment each frame into regions and, at each region we utilize static features and dynamic features computed from the low and middle levels to obtain saliency cues. By changing the scale in segmentation, we explore saliency cues in multiple scales. To keep temporal consistency across consecutive frames, we introduce adaptive temporal windows that are computed from motion of segmented regions. Fusing saliency cues obtained in multiscale using adaptive temporal windows allows us to obtain our spatiotemporal saliency map. Our intensive experiments using publicly available datasets demonstrate that our method outperforms the state-of-the-arts.  

The proposed method can detect salient objects from dynamic scenes, but it still cannot work well for complex dynamic background. This problem could be tackled by incorporating high-level knowledge to further improve the saliency maps. Our proposed method has flexibility in utilizing more features. Although the current method utilizes low-level and middle-level features, we plan to utilize high-level features such as human faces or semantic knowledge exploited from videos. We believe saliency maps generated by our proposed method can be further used for efficient object detection and action recognition.

\bibliography{MyBibtex}
\bibliographystyle{ws-ijprai}






\end{document}